%% file: main.tex
\documentclass[10pt,twocolumn,letterpaper]{article}

\usepackage[pagenumbers]{cvpr} 

\input{preamble}
\definecolor{cvprblue}{rgb}{0.21,0.49,0.74}
\usepackage[pagebackref,breaklinks,colorlinks,allcolors=cvprblue]{hyperref}
\usepackage{wrapfig,booktabs}
\usepackage{booktabs}  
\usepackage{multirow}  
\usepackage{amssymb}   
\usepackage{graphicx}
\usepackage{makecell}

\def\paperID{783} 
\def\confName{CVPR}
\def\confYear{2026}

\title{Are We Ready for RL in Text-to-3D Generation?\\ A Progressive Investigation}

\author{Yiwen Tang$^{1,4}$*, Zoey Guo$^{3}$*, Kaixin Zhu$^{2}$*, Ray Zhang$^{3}$*, Qizhi Chen$^{4}$, Dongzhi Jiang$^{3}$, \vspace{0.1cm}\\  Junli Liu$^{4}$, Bohan Zeng$^{2}$, Haoming Song$^{4}$, Delin Qu$^{4}$, Tianyi Bai$^{5}$, Dan Xu$^{5}$, Wentao Zhang$^{2}$, Bin Zhao$^{1,4}$\vspace{0.1cm}\\
$^{1}$Northwestern Polytechnical University  \quad $^{2}$Peking University\\ $^{3}$The Chinese University of Hong Kong \quad  $^{4}$Shanghai AI Lab\\   $^{5}$The Hong Kong University of Science and Technology 
}

\begin{document}
\maketitle
\input{sec/0_abstract}    
\input{sec/1_intro}
\input{sec/2_formatting}

\input{sec/3_finalcopy}

\loop\ifnum\value{page}<18
    \null\newpage
\repeat

{
    \small
    \bibliographystyle{ieeenat_fullname}
    \bibliography{main}
}
\end{document}

%% file: preamble.tex
\usepackage[utf8]{inputenc}

%% file: sec/0_abstract.tex
\begin{abstract}
Reinforcement learning (RL), earlier proven to be effective in large language and multi-modal models, has been successfully extended to enhance 2D image generation recently. However, applying RL to 3D generation remains largely unexplored due to the higher spatial complexity of 3D objects, which require globally consistent geometry and fine-grained local textures. This makes 3D generation significantly sensitive to reward designs and RL algorithms. To address these challenges, we conduct the first systematic study of RL for text-to-3D autoregressive generation across several dimensions. (1) Reward designs: We evaluate reward dimensions and model choices, showing that alignment with human preference is crucial, and that general multi-modal models provide robust signal for 3D attributes. (2) RL algorithms: We study GRPO variants, highlighting the effectiveness of token-level optimization, and further investigate the scaling of training data and iterations.
(3) Text-to-3D Benchmarks: Since existing benchmarks fail to measure implicit reasoning abilities in 3D generation models, we introduce MME-3DR. (4) Advanced RL paradigms: Motivated by the natural hierarchy of 3D generation, we propose Hi-GRPO, which optimizes the global-to-local hierarchical 3D generation through dedicated reward ensembles. Based on these insights, we develop AR3D-R1, the first RL-enhanced text-to-3D model, expert from coarse shape to texture refinement. We hope this study provides insights into RL-driven reasoning for 3D generation. Code is released at \url{https://github.com/Ivan-Tang-3D/3DGen-R1}.
\end{abstract}

%% file: sec/1_intro.tex
\section{Introduction}
\label{sec:intro}

Large Language Models (LLMs) and Large Multi-modal Models (LMMs) have achieved strong results in tasks like text generation, image grounding~\cite{li2024llava}, and video understanding~\cite{lin2024video}, yet still struggle with complex reasoning tasks such as mathematical problem solving~\cite{zhang2024mathverse} and code generation~\cite{seed2025seed}.
Recently, driven by Chain-of-Thought (CoT) reasoning capabilities that emerge through reinforcement learning (RL), advanced models such as OpenAI o3~\cite{openai2025} and DeepSeek-R1~\cite{guo2025deepseek} achieve significant gains on these challenging tasks.
As shown in Figure~\ref{teaser}, RL training has expanded beyond understanding tasks to multi-modal generation, particularly in autoregressive text-to-image models. The prior work Image Generation with CoT~\cite{guo2025can} demonstrated the effectiveness of Direct Preference Optimization (DPO)~\cite{rafailov2023direct} in improving intermediate generation processes. More recently, several studies~\cite{jiang2025t2i,tong2025delving} have explored the application of Group Relative Policy Optimization (GRPO)~\cite{shao2024deepseekmath} to 2D generation.
However, 3D autoregressive generation models~\cite{hao2024meshtron,siddiqui2024meshgpt} have primarily focused on pre-training and fine-tuning approaches.

This raises the question: \textit{\textbf{Can RL training be applied to text-to-3D generation, strengthening the step-by-step process of 3D autoregressive models?}}
While RL has shown promise in text-to-image generation, these strategies cannot be directly applied to text-to-3D generation.
3D assets involve coupled geometric and textural properties operating in higher spatial dimensionality, making RL training more sensitive to reward designs and algorithmic choices.
Moreover, 3D generation requires coherent joint optimization across multiple object components

Therefore, we systematically investigate the potential of the RL training for 3D autoregressive generation. Building upon the GRPO algorithm and the 3D discrete model ShapeLLM-Omni~\cite{ye2025shapellm}, and inspired by recent advances in 2D generation~\cite{jiang2025t2i}, we introduce a reasoning-guided framework where the model first generates textual reasoning that subsequently guides token-level 3D generation. We evaluate our approach on Toys4K~\cite{stojanov2021using}. As shown in the right part of Figure~\ref{teaser}, our analysis focuses on following perspectives:

\begin{figure*}[tb!]
  \centering
  \vspace{-0.4cm}
  \scalebox{0.95}{\includegraphics[width=\linewidth]{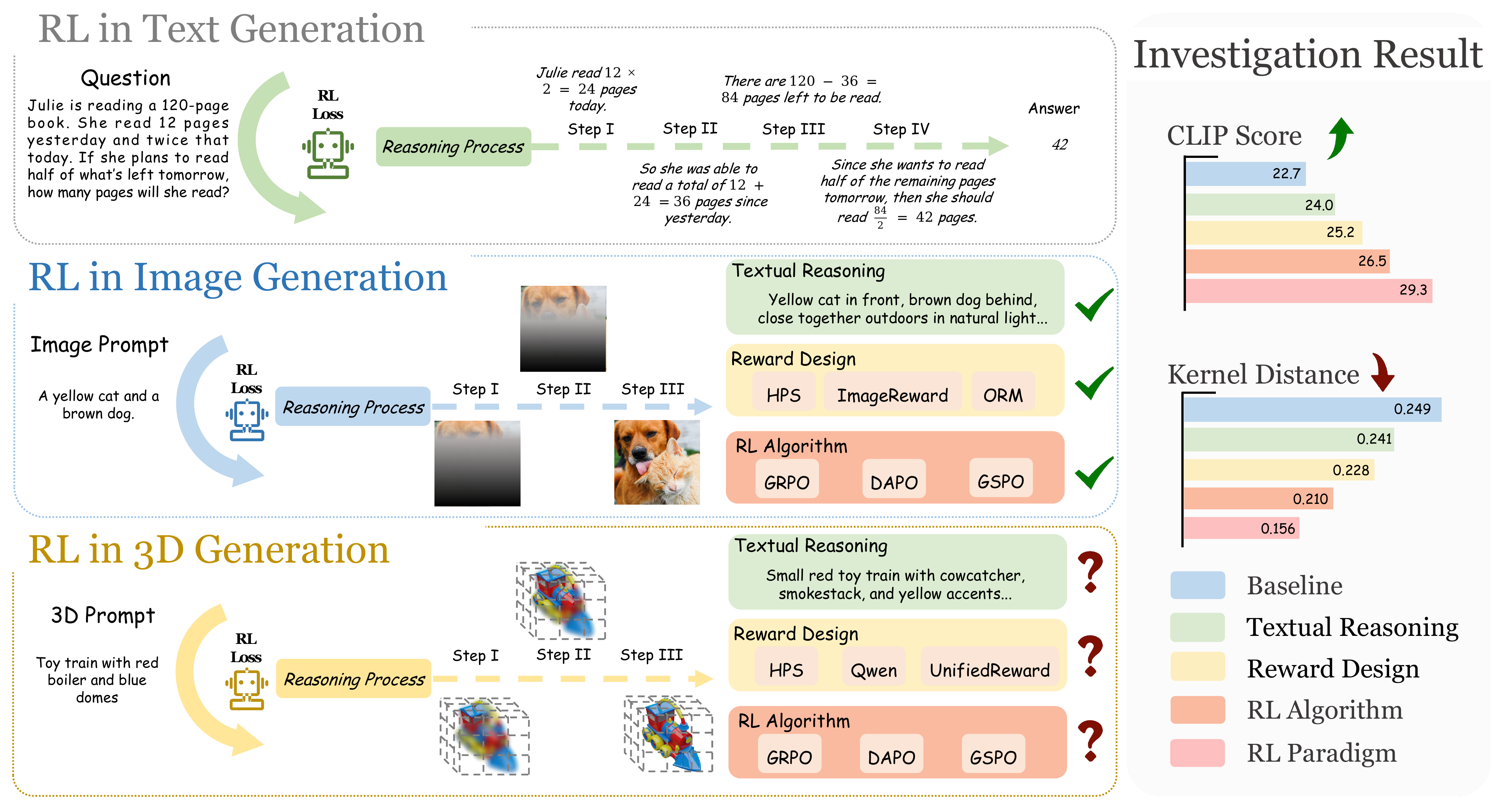}}
  \caption{\textbf{The Illustration of RL in Text, Image and 3D Generation Tasks.}
  Left: In text generation, RL induces textual reasoning, whereas in 2D and 3D autoregressive generation, RL primarily improves token-level generation. Although recent 2D studies have explored reasoning-guided generation, diverse reward models and RL algorithms, such approaches remain examined in 3D generation.
  Right: We present the effects of different strategies on RL performance, using ShapeLLM-Omni as the baseline model. KD is reported ×100.}
\label{teaser}
\end{figure*}

\textbullet\ \textbf{Impact of Different Reward Models.} 
In 3D generation, reward models serve diverse roles, including modeling human preference distributions, measuring prompt alignment, and enforcing multi-view consistency. These rewards vary from task-specific trained models to prompt-driven general LMMs.
Understanding how different reward sources shape policy behavior is critical for RL training.

\textbf{\textit{Observations.}}
\textit{
1) Functions of Reward Models.
Aligning with human preferences is crucial for 3D autoregressive generation. Additionally, enforcing consistency with text prompt and incorporating 3D aesthetic priors further enhance generation quality.
2) Forms of Reward Models.
Compared to specialized reward models, exclusive reliance on general LMMs for task-specific evaluation introduces systematic bias. However, LMMs surprisingly demonstrate strong robustness for 3D-relevant attributes.
}

\textbullet\ \textbf{Impact of Different RL Algorithms.}
Recent works have introduced GRPO variants~\cite{yu2025dapo,yang2025qwen3} to improve LLM reasoning. For instance, DAPO~\cite{yu2025dapo} enforces consistent token-level averaging in loss computation and promotes sequence diversity, whereas GSPO~\cite{yang2025qwen3} clips sequence-level likelihood differences between new and old policies to align with sequence-level rewards. We therefore ask whether these improvements also benefit 3D generation.

\textbf{\textit{Observations.}}
\textit{1) The RL loss of 3D autoregressive models benefits more from token-level averaging, as it better captures global structural differences during generation.
In contrast, sequence-level operations provide limited gains.
2) Simple techniques in DAPO, such as dynamic sampling, are sufficient to stabilize training for text-to-3D generation.
3) Data scaling effectively improves performance, whereas iteration scaling demands careful calibration.}

\textbullet\ \textbf{Exploration of Text-to-3D Benchmarks.}
Current text-to-3D benchmarks fail to evaluate models under reasoning-heavy conditions. While models perform well on simple prompts, we observe consistent failures across five categories: (1) Spatial \& structural geometry, (2) Mechanical affordances, (3) Biological \& organic shapes, (4) World-knowledge rare objects, and (5) Stylized representation. As a result, existing benchmarks overestimate generation models and ignore their intrinsic reasoning abilities.
To bridge this gap, we propose MME-3DR, the first benchmark designed for these reasoning-intensive 3D cases. It contains 249 annotated 3D objects spanning the five challenging categories. Experiments on several text-to-3D models validate the effectiveness of MME-3DR.

\textbf{\textit{Observations.}}
\textit{1)Recent text-to-3D models demonstrate reasonable performance on biological objects and those with well-defined mechanical structures, yet they remain fragile across other categories. 
2) After RL training, the model achieves substantial improvements across all five categories compared to the base model.
3) MME-3DR serves dual purposes: measuring generation quality and evaluating implicit reasoning capabilities.}

\begin{figure*}[tb!]
  \centering
  \scalebox{0.95}{\includegraphics[width=\linewidth]{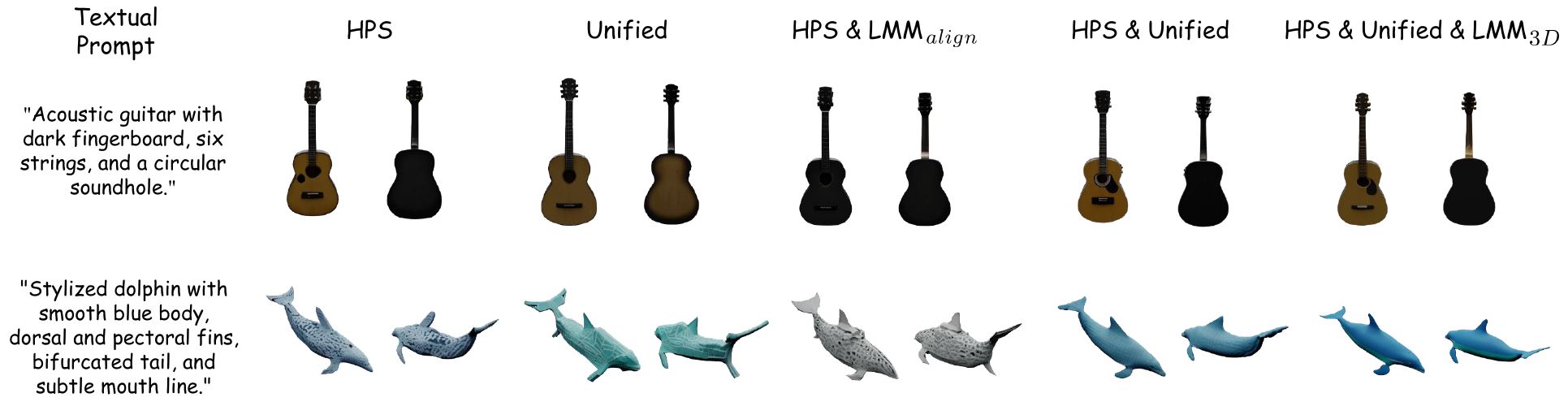}}
  \caption{\textbf{Visualization Results of the Impact of Different Reward Models.}}
\label{diffreward}
\end{figure*}

\textbullet\ \textbf{Exploration of RL Paradigms.}
During training, we observe that the model first constructs the global geometry and then progressively refines local textures in later stages, resembling human 3D perception process. This indicates that RL paradigms leveraging textual reasoning for direct 3D guidance can be further enhanced. To jointly optimize the hierarchical 3D generation within a single iteration, we introduce Hi-GRPO.
With unified generation and understanding abilities of ShapeLLM-Omni, our method first prompts the model to plan the global structure and produce high-level semantic reasoning for token-level generation, yielding a coarse 3D shape that captures geometry but lacks texture details. 
In the second step, we feed both the first-step CoT and the original prompt into the model to obtain low-level visual reasoning and generate a texture-refined 3D object.
In each iteration, we sequentially generate multiple coarse shapes and corresponding refined models for each prompt, and introduce two specialized ensembles of expert reward models to compute group-relative rewards for both steps. Building on these strategies, we develop AR3D-R1, the first RL---enhanced 3D autoregressive model.

\textbf{\textit{Observations.}}
\textit{1) AR3D-R1 demonstrates a coarse-to-fine progression during inference, evolving from rough shapes to detailed textures. This behavior aligns with our training procedure and validates the effectiveness of Hi-GRPO.
2) AR3D-R1 exhibits strong reasoning capability on MME-3DR and outperforms Trellis on benchmarks.}

In summary, our core contributions are as follows:

\textbullet\ We are the first to systematically introduce reinforcement learning into text-to-3D autoregressive generation, conducting an in-depth analysis from multiple perspectives.

\textbullet\ By examining reward model design and RL algorithm selection, we show that both must be tailored to 3D domain knowledge and that appropriate reward models significantly enhance overall performance.

\textbullet\ From the perspective of text-to-3D benchmark, we observe that existing benchmarks focus on object diversity while neglecting evaluation of model capability, and therefore introduce MME-3DR to assess the intrinsic reasoning abilities of 3D generation models.

\textbullet\ From the RL paradigm perspective, we reveal an inherent hierarchy in 3D generation---from coherent geometry to texture-refined object, and propose Hi-GRPO, an advanced RL paradigm that jointly optimizes both steps in a single iteration. Building on these, we develop AR3D-R1, which outperforms current text-to-3D models.

\section{Related Work}

\subsection{RL for LLM}

Advanced LLMs such as OpenAI o3~\cite{openai2025} and DeepSeek-R1~\cite{guo2025deepseek} have demonstrated strong reasoning abilities by combining Chain-of-Thought (CoT) reasoning with reinforcement learning (RL). DeepSeek-R1 introduces rule-based rewards and GRPO~\cite{shao2024deepseekmath}, enabling models to conduct extensive internal reasoning before producing an answer, with rewards guiding correctness and format compliance. This paradigm has also been extended to multimodal LLMs~\cite{shen2025vlm,zheng2025deepeyes,huang2025vision,feng2025video}, where RL is adapted for visual understanding by jointly processing images and text for step-by-step reasoning. These RL-driven approaches have proven effective across mathematical problem-solving~\cite{zhang2024mathverse,shao2024deepseekmath} and code generation~\cite{seed2025seed}, establishing RL as a key technique for eliciting advanced capabilities in large-scale models.

\subsection{RL for 2D Generation}

RL has also been effectively applied to text-to-image generation. Image-Generation-CoT~\cite{guo2025can} first frames progressive image token generation as a reasoning process and applies DPO~\cite{rafailov2023direct} accordingly. T2I-R1~\cite{jiang2025t2i} extends this idea by distinguishing two levels of CoT—semantic-level planning and token-level patch generation—and introduces BiCoT-GRPO to jointly optimize both using an ensemble of vision experts as reward models. Recent work~\cite{tong2025delving} comparing DPO and GRPO shows that GRPO offers better text–image alignment and aesthetic quality through group-relative policy updates. Together, these studies highlight that well-designed sequence-level rewards and multi-dimensional evaluation are essential for producing semantically consistent and visually appealing images in autoregressive models.
For diffusion models, Dance-GRPO~\cite{xue2025dancegrpo} introduces a stepwise, motion-aware reward that aligns policy updates with temporal dynamics, enabling more coherent and physically plausible generation. Flow-GRPO~\cite{liu2025flow} extends GRPO to flow-matching models by coupling policy optimization with flow objectives, yielding smoother training and improved stability. These methods show that RL can effectively enhance controllability and consistency in diffusion and flow-based generative models.

\subsection{Text-to-3D Generation}

Text-to-3D generation has progressed from two-stage pipelines~\cite{yang2024hunyuan3d,xu2023dream3d} to native diffusion models~\cite{xiang2025structured,chen20253dtopia} and, more recently, autoregressive approaches~\cite{siddiqui2024meshgpt,chen2024meshanything,zhao2025deepmesh,ye2025shapellm}. 
Two-stage methods, like Dream3D~\cite{xu2023dream3d}, first generate a high-quality 3D shape prior from text using a text-to-image diffusion model and then refine it as a neural radiance field, but this pipeline suffers from error accumulation between stages and limited 3D consistency inherited from the 2D diffusion backbone.
Native diffusion models, like Trellis~\cite{xiang2025structured}, leverage structured 3D latent representations to directly generate high-fidelity 3D content, but their strong performance comes at the cost of significant computational demands.
Autoregressive models alleviate these limitations by discretizing 3D content into token sequences. MeshGPT~\cite{siddiqui2024meshgpt} uses decoder-only transformer to model triangle meshes as sequences, and MeshAnything~\cite{chen2024meshanything} demonstrates scalable artist-grade mesh generation using autoregressive transformers. DeepMesh~\cite{zhao2025deepmesh} provides an early attempt to incorporate DPO~\cite{rafailov2023direct} into autoregressive 3D creation, and LLaMA-Mesh~\cite{wang2024llama} represents 3D OBJ files as text to unify language and 3D representations.
ShapeLLM-Omni~\cite{ye2025shapellm} proposes a unified multimodal LLM for 3D generation and understanding by discretizing 3D shapes into tokens with a 3D VQVAE. The model follows a text→voxel pipeline, where the LLM predicts discrete 3D latent tokens that are decoded by the VQVAE into voxel grids, which are then further converted into meshes using Rectified Flow
model~\cite{xiang2025structured} for rendering. This design enables a single LLM to support text-to-3D generation, 3D understanding, and editing within one coherent framework.
Despite these developments, RL training for 3D autoregressive models remains largely unexplored. In contrast to 2D generation, where RL has shown clear benefits, 3D generation introduces additional challenges—greater spatial complexity, stricter global geometry constraints, and fine-grained local details—making it more sensitive to reward design and optimization choices. These factors highlight the need for systematic RL strategies tailored to text-to-3D generation.

%% file: sec/2_formatting.tex
\section{Preliminary}

\textbf{3D Autoregressive Generation.} 
Compared with two-stage pipelines~\cite{xu2023dream3d} and native 3D diffusion models~\cite{yang2024hunyuan3d,zhao2025hunyuan3d,wu2025direct3d}, 3D autoregressive generation models~\cite{zhao2025deepmesh,siddiqui2024meshgpt,wang2024crm,chen2024meshanything} directly discretize 3D objects into token sequences. This can be achieved by compressing and quantizing 3D shapes with VQVAE~\cite{ye2025shapellm, chen2025sar3d}, or by applying mesh tokenization techniques~\cite{zhao2025deepmesh, chen2025meshanything} to discretize vertices and faces. 
Furthermore, LLaMA-Mesh~\cite{wang2024llama} treats 3D OBJ files as plain text, utilizing natural language as the interface for mesh generation and understanding. Building further, ShapeLLM-Omni~\cite{ye2025shapellm} integrates Qwen2.5-VL with a 3D VQVAE module to unify 3D generation and understanding, enabling autoregressive prediction over discrete 3D tokens and text. 
More details can be found in the supplementary material.

\begin{table}[tb!]
\centering
\caption{\textbf{Quantitative comparisons using Toys4k for Different Reward Models.} 
HPS refers to HPS v2.1, which outputs human preference rewards. 
Unified denotes UnifiedReward-2.0-Qwen7B, which jointly evaluates aesthetic quality and prompt alignment. 
LMM$_{\text{align}}$ employs Qwen2.5-VL-7B to replace UnifiedReward functionality. 
LMM$_{\text{3D}}$ utilizes Qwen2.5-VL to assess 3D consistency. (KD is reported ×100)}
\label{tab:reward_comparison}
\scalebox{0.8}{
\begin{tabular}{ccccc@{\hspace{1em}}c}
\toprule
\multicolumn{4}{c}{Reward Model} & \multicolumn{2}{c}{Metrics} \\
\cmidrule(r){1-4} \cmidrule(l){5-6}
HPS & Unified & LMM$_{\text{align}}$ & LMM$_{\text{3D}}$ & CLIP Score$\uparrow$ & KD$_{\text{incep}}$$\downarrow$ \\
\midrule
- & - & - & - & 22.7 & 0.249 \\
\checkmark & - & - & - & 24.0 & 0.241 \\
- & \checkmark & - & - & 23.5 & 0.246 \\
- & - & - & \checkmark & 23.3 & 0.245 \\
\checkmark & \checkmark & - & - & 24.6 & 0.235 \\
\checkmark & - & \checkmark & - & 24.2 & 0.238 \\
\checkmark & \checkmark & - & \checkmark & \textbf{25.2} & \textbf{0.228} \\
\bottomrule
\end{tabular}
}
\end{table}

\begin{figure*}[tb!]
\centering
\begin{minipage}{0.62\linewidth}
\centering
\captionof{table}{\textbf{Quantitative comparisons using Toys4k for Different RL algorithms.} In DAPO, Clip, Sampling, Token Avg., and KL Remov. correspond to Decoupled Clip, Dynamic Sampling, Token-level Loss Aggregation, and KL Penalty Removal, respectively. For GSPO, Seq. Opt. indicates that both importance sampling and clipping are performed at the sequence level. (KD is reported ×100)}
\label{tab:rlC_omp}
\scalebox{0.85}{
\begin{tabular}{cccccc@{\hspace{1em}}c}
\toprule
\multicolumn{4}{c}{DAPO} & \multicolumn{1}{c}{GSPO} & \multicolumn{2}{c}{Metrics} \\
\cmidrule(r){1-4} \cmidrule(r){5-5} \cmidrule(l){6-7}
Clip & Sampling & Token Avg. & KL Remov. & Seq. Opt. & CLIP Score$\uparrow$ & KD$_{\text{incep}}$$\downarrow$ \\
\midrule
- & - & - & -  & - & 25.2 & 0.228 \\
- & \checkmark & - & -  & - & 25.8 & 0.219 \\
- & \checkmark & - & -  & \checkmark & 25.5 & 0.223 \\
- & \checkmark & \checkmark  & - & - & 26.3 & 0.214 \\
- & \checkmark & \checkmark & \checkmark  & - & 25.9 & 0.213 \\
\checkmark & \checkmark & \checkmark & -  & - & \textbf{26.5} & \textbf{0.210} \\
\bottomrule
\end{tabular}
\vspace{-0.4cm}
}
\end{minipage}
\hspace{0.2cm}
\begin{minipage}{0.32\linewidth}
\centering
\includegraphics[width=0.9\linewidth]{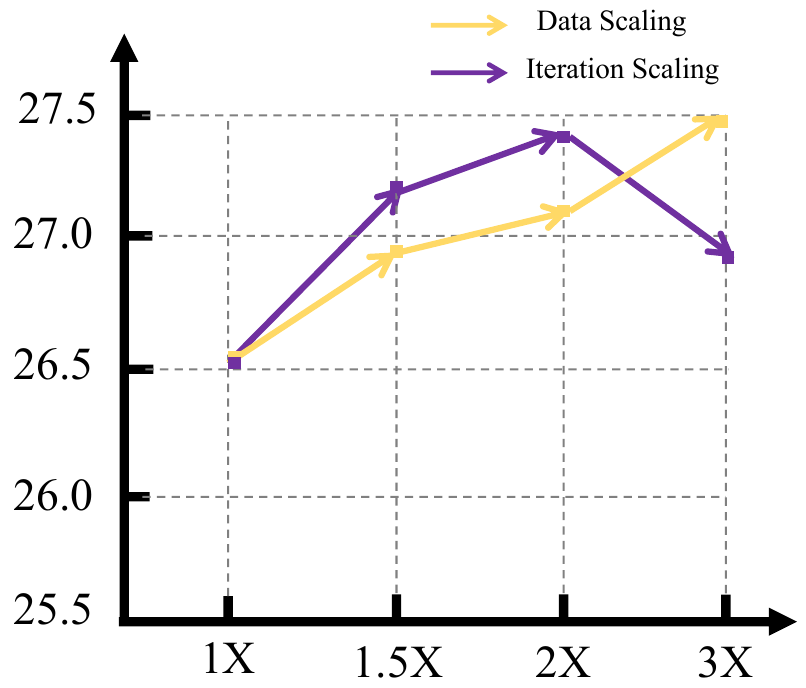}
\caption{\textbf{Effects of Scaling Strategies.}
We examine the effects of data scaling and training iteration scaling strategies on clip score.}
\label{scaling}
\end{minipage}
\end{figure*}

\textbf{RL Algorithm.}
GRPO~\cite{shao2024deepseekmath} is an on-policy reinforcement learning algorithm that enhances PPO~\cite{schulman2017proximal} by removing the value function and using group-wise reward comparisons.
For a given prompt, $G$ responses $\{o_i\}_{i=1}^{G}$ are sampled from the old policy $\pi_{\theta_{\text{old}}}$.
Each response receives a reward $R_i$ from the reward model, and the advantage is computed by normalizing rewards within the group:
\[
A_i = \frac{R_i - \text{mean}(\{R_i\}_{i=1}^{G})}{\text{std}(\{R_i\}_{i=1}^{G})}.
\]
GRPO uses PPO-like clipping and introduces a KL penalty term between the policy $\pi_{\theta}$ and the reference $\pi_{\theta_{\text{ref}}}$.

In text reasoning tasks, rewards are depend on the correctness of the final answer and the output format.

\begin{table}[tb!]
\centering
\caption{\textbf{Effectiveness of textual reasoning.} We employ HPS V2.1 as the reward and adopt GRPO.}
\label{tab:reward-types}
\scalebox{0.8}{ 
\begin{tabular}{lc}
\toprule
Textual Reasoning & CLIP Score \\
\midrule
Base Model  & 22.7      \\
W/O & 23.4    \\
W/ & 24.0    \\
\bottomrule
\end{tabular}
}
\end{table}

\textbf{Experimental Settings.}
We adopt ShapeLLM-Omni as our baseline, a recent state-of-the-art 3D autoregressive model. 
We curate 8,400 short captions from 3D object datasets as training prompts~\cite{collins2022abo, deitke2023objaverse,khanna2024habitat}.
To systematically evaluate different strategies, we randomly select 800 samples from Toys4K as our test set.
Given ShapeLLM-Omni understanding-and-generation capability and the recent advances in 2D generation~\cite{jiang2025t2i}, we do not directly generate 3D objects. 

Instead, each training iteration begins by prompting the model to imagine the object and produce $G$ textual descriptions, followed by generating one 3D object conditioned on each description, where $G = 8$. Table~\ref{tab:reward-types} shows that textual reasoning prior to 3D token-level generation yields greater RL potential than direct generation.
In the following sections, we explore the applicability of GRPO to text-to-3D generation from four perspectives: Reward models (Sec.~\ref{sec:reward}), RL algorithm choice (Sec.~\ref{sec:rl_agl}), Text-to-3D benchmarks (Sec.~\ref{sec:benchmark}) and the advanced RL paradigm design (Sec.~\ref{paradigm}).

\section{Impact of Different Reward Models}
\label{sec:reward}

Reinforcement learning has been proven effective for large language models and 2D generation, using reward models to capture human preferences and improve aesthetic quality. However, RL for 3D autoregressive generation remains underexplored. We therefore first study how reward model capability affects RL performance using GRPO with group size $G=8$ and one policy update per iteration.

\textbf{Reward Model Design.}
Unlike code or math tasks, where deterministic verification functions provide direct rewards, multi-modal generation requires learned reward models. 3D generation faces unique challenges compared to 2D: (1) Complex design: 3D objects lack canonical viewpoints, requiring multi-dimensional reward systems that jointly evaluate realism, semantic alignment, and structural integrity; (2) Multi-view assessment: Evaluation must ensure cross-view consistency, verifying that shapes and textures across viewpoints form structurally valid objects.
We therefore investigate reward model paradigms across different evaluation dimensions and their combination strategies.

\textbullet\ \textbf{Human Preference.} 
Human preference models, such as the HPS series~\cite{wu2023human}, are vision–language models trained on large-scale human-annotated image-ranking datasets.
The model takes a prompt and multiple rendered views as input, assigns a score to each view, and uses the highest score to represent the overall 3D object visual quality.

\textbullet\ \textbf{Prompt Alignment \& Aesthetic Quality.}
Specialized reward models, such as UnifiedReward~\cite{wang2025unified}, evaluate each rendered view of a 3D object with three scores: (1) Prompt---image alignment, (2) Image logical coherence, and (3) Image style appealing. These scores are summed, and the maximum score across views is taken as the final reward. The latter two terms capture aesthetic quality. 
In contrast, general LMMs, like Qwen2.5-VL, jointly process the prompt and all views to generate a single reasoning-based score, used for either alignment or aesthetics.

\textbullet\ \textbf{3D Consistency.} 
There is no reward model trained specifically for 3D consistency. However, we observe that advanced LMMs, such as Qwen2.5-VL~\cite{bai2025qwen2}, exhibit strong 3D understanding and can assess cross-view spatial consistency. The model evaluates consistency across three dimensions: (1) shape outline across views, (2) appearance, and (3) object parts. Each dimension is rated from 0 to 1, and their sum is used as the overall 3D consistency score.

\begin{figure*}[tb!]
  \centering
  \includegraphics[scale=0.22]{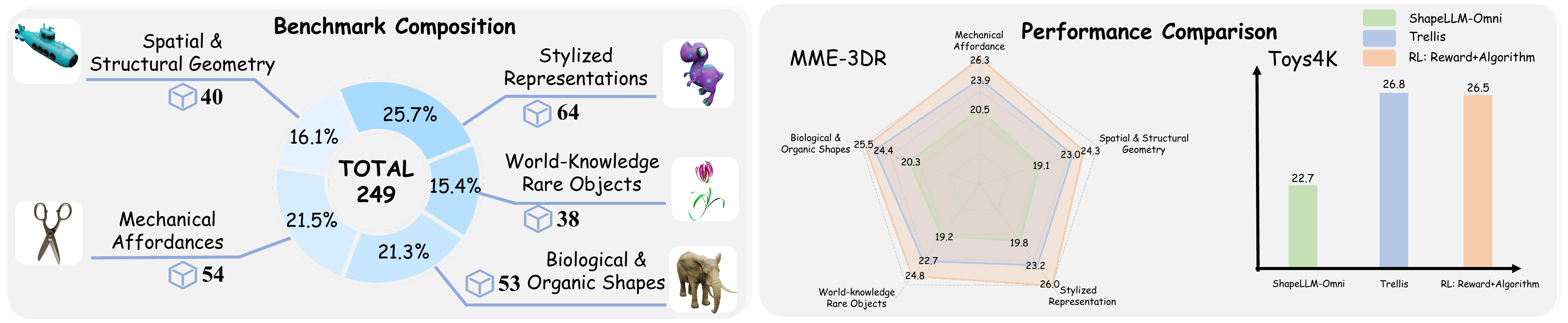}
  \caption{\textbf{MME-3DR Benchmark Analysis.} Left: MME-3DR contains 249 complex 3D objects across five categories. Right: We compare the performance of ShapeLLM-Omni, the SOTA model Trellis, and our RL-enhanced model on MME-3DR and Toys4K. Using CLIP Score as the metric, the results highlight the importance of implicit reasoning ability.}
\label{benchmark}
\end{figure*}

\textbf{Experimental Analysis and Insights.} 
Detailed results are reported in Table~\ref{tab:reward_comparison}, and qualitative visualizations are shown in Figure~\ref{diffreward}. 
We standardize the reward evaluation by sampling six rendered views for each 3D-generated object.
Based on these results, we draw two key insights:

\textbullet\ \textit{Human Preference reward serves as the core signal for RL in 3D autoregressive generation. Other reward dimensions offer limited standalone benefit but consistently improve performance when added on top of the preference reward.} As shown in Table~\ref{tab:reward_comparison}, HPS V2.1 delivers the strongest gains among single-reward settings, and combining it with UnifiedReward or Qwen2.5-VL yields up to an additional 0.6 performance improvement.

\textbullet\ \textit{For a given reward dimension, specialized reward models show greater robustness than LMMs. However, for multi-view objectives such as 3D consistency, LMMs exhibit superior generalization.} As shown in Table~\ref{tab:reward_comparison}, combining HPS V2.1 with UnifiedReward outperforms pairing it with Qwen2.5-VL by 0.4. In contrast, when Qwen2.5-VL is used to assess 3D consistency, it delivers a 0.6 improvement in CLIP score.
As shown in Figure~\ref{diffreward}, the 3D consistency reward effectively enhances coherence in color, texture, and geometry for both the guitar and the dolphin.

\section{Impact of Different RL Algorithms}
\label{sec:rl_agl}

GRPO~\cite{shao2024deepseekmath}, a widely used on-policy RL algorithm, has been applied to LLMs, LMMs, and 2D generation to enhance reasoning capabilities and generalization. Recentl GRPO variants, such as DAPO~\cite{yu2025dapo} and GSPO~\cite{yang2025qwen3}, have emerged, demonstrating superior efficiency and effectiveness on mathematical and coding tasks. However, their application to generation tasks remains underexplored. We therefore train and evaluate GRPO, DAPO, and GSPO on 3D autoregressive generation to systematically assess their respective advantages.

\textbf{DAPO.} 
DAPO~\cite{yu2025dapo} mitigates entropy collapse and training instability in vanilla GRPO by introducing several techniques, some of which are promising for 3D generation: (1) decoupled clipping bounds to enhance exploration and avoid oversmoothing of 3D object, (2) dynamic sample filtering to focus on medium-complexity 3D cases, (3) token-level loss aggregation to reduce bias toward trivial shapes, and (4) removal of KL regularization for more flexible policy updates. We adopt the same settings as GRPO, with a group size of 8 and one iteration per update.

\textbf{GSPO.} To mitigate expert-activation fluctuations from token-level optimization in GRPO~\cite{zheng2025group}, GSPO~\cite{yang2025qwen3} shifts optimization to the sequence level. It performs importance sampling and clipping based on sequence likelihood under the current and reference policy models. This ensures that each 3D object is optimized as a coherent whole, preventing local token-level conflicts that could lead to inconsistent geometry. We adopt the same training setting as GRPO.

\textbf{Experimental Analysis and Insights.} 
As shown in Table~\ref{tab:rlC_omp}, we systematically analyze RL algorithms---GRPO, DAPO, and GSPO---for 3D autoregressive generation, examining their strengths, limitations, and effective combinations. We further validate scaling strategies across training data and training iterations based on the optimal RL configuration in Figure~\ref{scaling}. The key findings are as follows:

\textbullet\ \textit{Compared to sequence-level operations, RL for 3D autoregressive generation favors token-level strategies.} As shown in Table~\ref{tab:rlC_omp}, under identical reward model settings, token-level averaging yields much larger gains than the sequence-level importance sampling and clipping.

\textbullet\ \textit{Simple techniques can already stabilize training, particularly Dynamic Sampling, as long as policy updates remain properly constrained.} As shown in Table~\ref{tab:rlC_omp}, Dynamic Sampling improves vanilla GRPO by 0.6. However, completely removing the KL penalty leads to a 0.4 drop, while a more controlled method such as Decoupled Clip still yields gains by encouraging low-probability token exploration.

\textbullet\ \textit{Scaling training data effectively mitigates preference bias and improves performance, while moderate iteration increases optimize results—though excessive training risks generalization degradation.}
As shown in Figure~\ref{scaling}, expanding the dataset by $1.5\times$, $2\times$, and $3\times$ yields gains of 0.4, 0.2, and 0.4, respectively. Doubling training iterations improves performance by 0.9, yet tripling them causes performance decline. This indicates significant generalization deterioration, likely attributable to overfitting on preference features.

\begin{figure*}[tb!]
  \centering
  \scalebox{0.9}{\includegraphics[width=\linewidth]{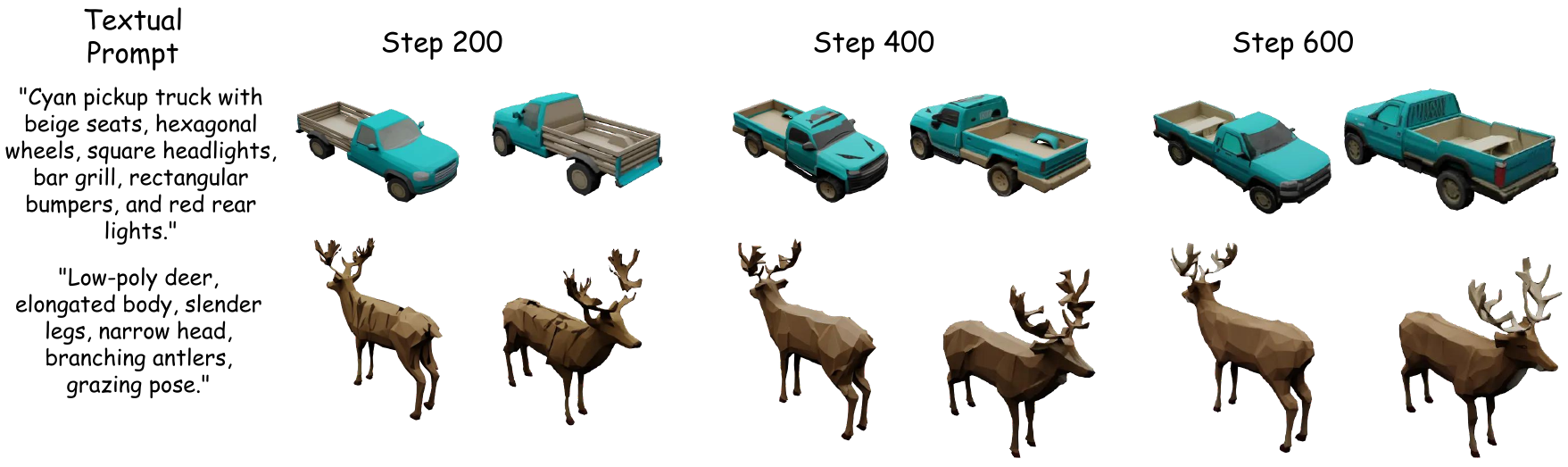}}
  \caption{\textbf{Visualization results across different training stages}}
\label{Step_compare}
\end{figure*}

%% file: sec/3_finalcopy.tex
\section{Exploration of Text-to-3D Benchmarks}
\label{sec:benchmark}

Current text-to-3D models generalize well on simple prompts, yet remain vulnerable to certain categories---a phenomenon observed in both autoregressive and diffusion models. As shown in Figure~\ref{benchmark}, these challenging cases mainly fall into five types:
\begin{enumerate}
    \item \textbf{Spatial \& Structural Geometry}: objects with complex spatial layouts and component arrangements;
    \item \textbf{Mechanical Affordances}: objects involving physical functionality or interactive mechanical components;
    \item \textbf{Biological \& Organic Shapes}: organisms (e.g., animals, plants) with dynamic organic characteristics;
    \item \textbf{World-Knowledge Rare Objects}: low-frequency concepts requiring broader real-world knowledge;
    \item \textbf{Stylized Representation}: non-photorealistic forms, , including cartoon, abstract, or stylistic interpretations.
\end{enumerate}
These types correspond to five core implicit reasoning abilities essential for 3D generation: spatial, physical, dynamic, knowledge-based, and abstract reasoning. The absence of these abilities reveals that current models depend heavily on memorization rather than genuine 3D understanding.
To address this gap, we propose MME-3DR, the first benchmark designed to evaluate the implicit reasoning capabilities of text-to-3D generation models.

\textbf{MME-3DR.}
We curate 249 complex 3D objects across five categories, carefully selected from Toys4K~\cite{stojanov2021using}, which are used neither in previous text-to-3D model training nor in current RL training. As shown in Figure~\ref{benchmark}, the benchmark comprises 16.1\% objects with complex spatial structures, 21.5\% with explicit mechanical and interactive components, 21.3\% non-rigid dynamic objects such as animals and plants, 15.4\% rare conceptual objects (e.g., fine-grained flower species), and 25.7\% stylized or non-realistic objects originating from artistic or toy-like designs.
Compared with our previous setup---randomly sampling 800 objects from the 105-category Toys4K dataset---MME-3DR intentionally balances samples across reasoning types critical for 3D generation, while maintaining broad object diversity.

\textbf{Experimental Analysis and Insights.} 
We evaluate ShapeLLM-Omni~\cite{ye2025shapellm}, Trellis~\cite{xiang2025structured}, and our RL-enhanced model on the proposed MME-3DR benchmark and the randomly sampled subset of Toys4K~\cite{stojanov2021using}. As shown in the right panel of Figure ~\ref{benchmark}, our key findings are as follows:

\textbullet\ \textit{Recent text-to-3D models perform reasonably on mechanical structures and non-rigid biological objects but struggle on the other three categories. 
RL training achieves substantial improvements across all five types.} 
Figure~\ref{benchmark} (radar chart) shows ShapeLLM-Omni and Trellis leading by over 1 point on mechanical affordances and biological \& organic shapes, likely due to higher training data prevalence.
RL training improves ShapeLLM-Omni by 5-6 points overall, with particularly notable gains in stylized representation, driven by enhanced implicit reasoning capabilities.

\textbullet\ \textit{MME-3DR evaluates implicit reasoning while simultaneously assessing general generation capability.} The bar chart in Figure~\ref{benchmark} reveals that Trellis significantly outperformes ShapeLLM-Omni on the randomly sampled Toys4K test set. This performance gap persists in MME-3DR, validating the effectiveness of its diverse object coverage.

\begin{figure*}[tb!]
  \centering
  \scalebox{0.93}{\includegraphics[width=\linewidth]{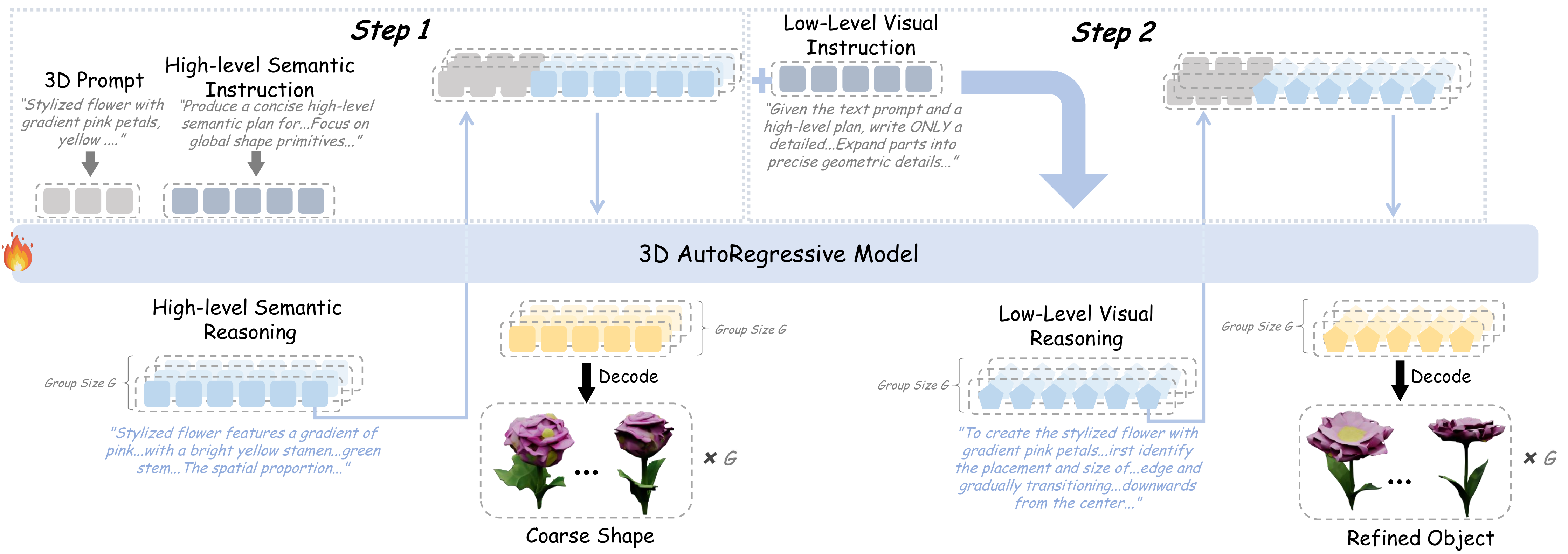}}
  \caption{\textbf{Framework of Hi-GRPO.}
  In Step 1, we instruct the model to generate high-level semantic reasoning based on the 3D prompt, and use it together with the prompt to produce a coarse 3D shape.
In Step 2, conditioned on the 3D prompt and the high-level semantic CoT, the model generates low-level visual reasoning focused on local appearance details, which is used to produce the refined 3D object.}
\label{pipeline}
\end{figure*}

\section{Exploration of RL Paradigms}
\label{paradigm}

In the text-to-3D task, we construct an RL paradigm for reasoning-guided 3D generation by leveraging the capabilities of ShapeLLM-Omni in text generation and token-wise 3D generation. Given a 3D textual prompt, the model first performs semantic reasoning to clarify user intent and resolve ambiguities. It then jointly models global structure and local texture details conditioned on both the prompt and the inferred reasoning, ultimately generating 3D tokens that are decoded into the final mesh.

While effective, the RL paradigm leaves substantial room for improvement. We observe that in early training stage, the model focuses on global geometry, producing coarse shapes with limited texture fidelity. As training progresses, reward signals drive refinement of materials and fine-grained textures, leading to clear gains in aesthetic quality and alignment with human preference. 
As shown in Figure~\ref{Step_compare}, we evaluate checkpoints at steps 200, 400, and 600 on identical prompts. Early-stage outputs resemble only a rough cyan pickup-truck-like shape; later, features such as beige seats, square headlights, rectangular bumpers, and red taillights gradually emerge. A similar trend is observed for the deer example: the antlers are largely absent early on, and subsequently evolve into well-defined, branched structures. 
This coarse-to-fine progression intuitively aligns with human 3D perception, where global geometry is recognized first, followed by fine-grained visual cues.

This raises a question: \textit{\textbf{Can hierarchical 3D generation process be integrated into RL paradigm to better align with the intrinsic nature of text-to-3D generation?}}
To this end, we introduce \textbf{Hi-GRPO}, which disentangles RL training into hierarchical coarse-to-fine steps. 
In each iteration, the model first predicts global structure and then refines local textures and details, producing high-fidelity 3D assets.

\begin{figure}[tb!]
  \centering
  \scalebox{0.85}{\includegraphics[width=\linewidth]{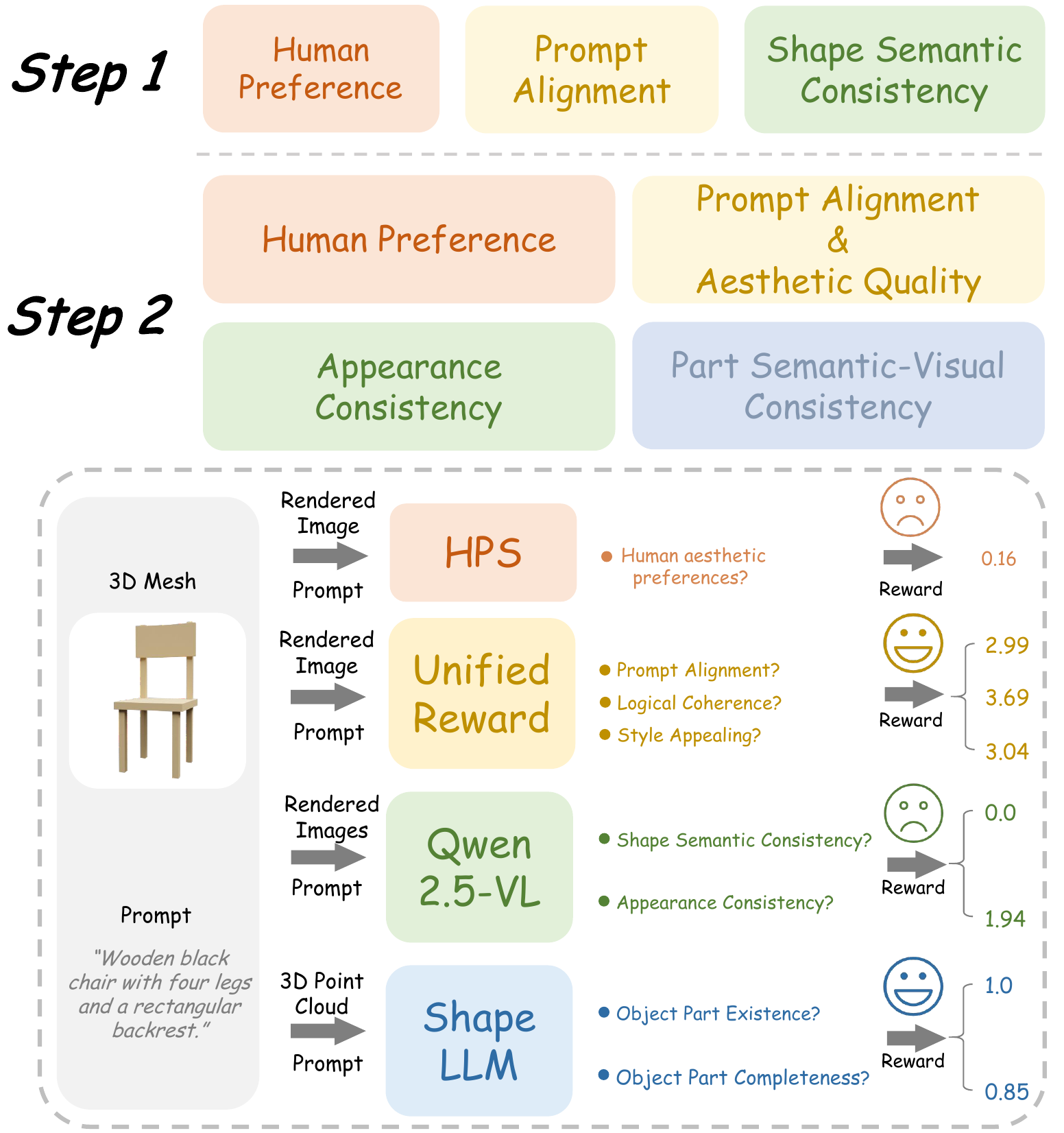}}
  \caption{\textbf{Illustration of the Reward Ensemble Design.}
  We design reward ensembles for steps in Hi-GRPO: step 1 focuses on global alignment, while step 2 emphasizes local refinement.}
  \label{reward_ensemble}
\end{figure}

\begin{figure*}[tb!]
  \centering
  \scalebox{0.8}{\includegraphics[width=\linewidth]{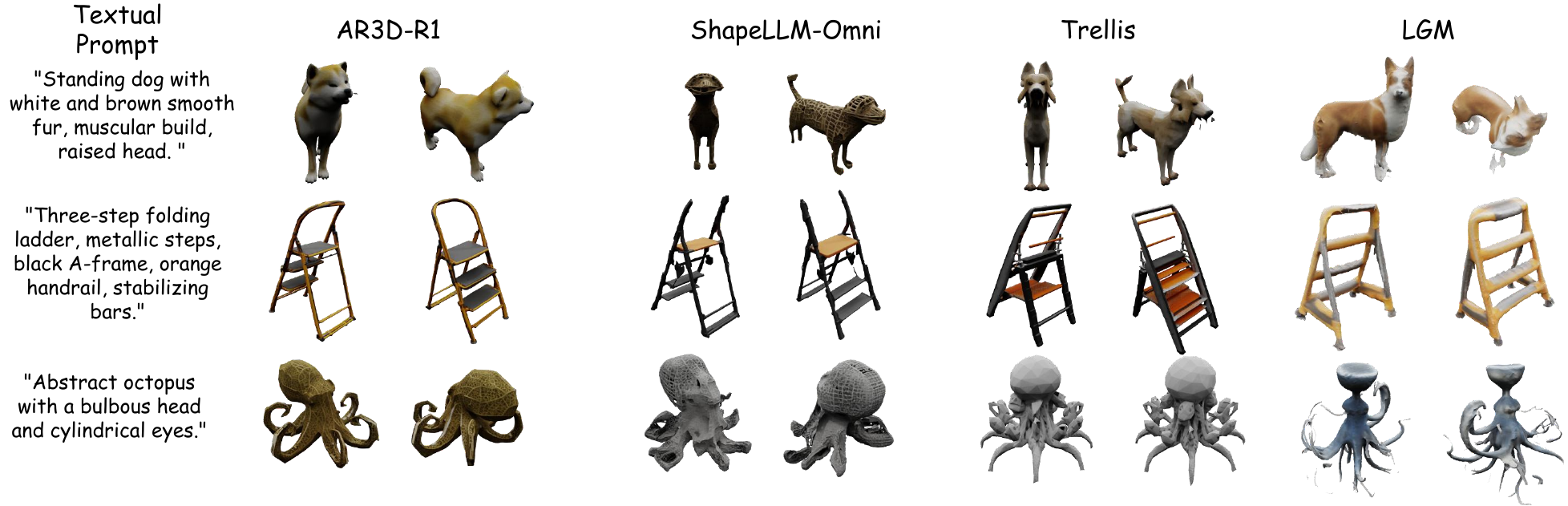}}
  \caption{\textbf{Qualitative Comparison of Text-to-3D Models.}}
\label{compare}
\end{figure*}

\begin{figure}[tb!]
    \centering
    \includegraphics[width=\linewidth]{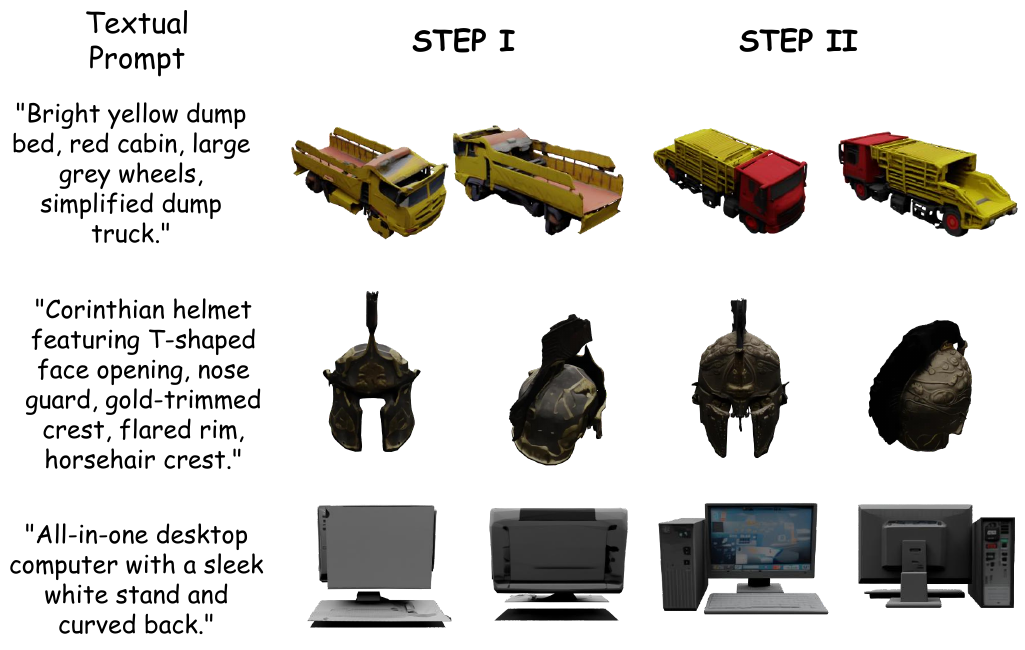}
    \caption{\textbf{Results of Different Steps during Inference.}}
    \label{diffstage}
\end{figure}

\begin{table*}[t!]
    \centering
    \captionof{table}{\textbf{Quantitative Comparison on Text-to-3D Generation Benchmarks.} (KD is reported $\times$100. $\dagger$: evaluated using shaded images of PBR meshes.)}
    \label{tab:performance}
    \scalebox{0.65}{
    \begin{tabular}{l|cccc|cccc}
    \toprule
    \multirow{2}{*}{Method} & \multicolumn{4}{c|}{MME-3DR} & \multicolumn{4}{c}{Toys4K} \\
    \cmidrule(lr){2-5}\cmidrule(lr){6-9}
     & CLIP$\uparrow$ & KD$_{\text{incep}}\downarrow$ &  KD$_{\text{dinov2}}\downarrow$ & FD$_{\text{incep}}\downarrow$
     & CLIP$\uparrow$ & KD$_{\text{incep}}\downarrow$ &  KD$_{\text{dinov2}}\downarrow$  & FD$_{\text{incep}}\downarrow$ \\
    \midrule
    LGM~\cite{tang2024lgm} & 16.3 & 1.507  & 49.10 & 47.8 & 20.6 & 1.192 & 36.79 & 41.0 \\
    3DTopia-XL~\cite{chen20253dtopia} & 15.9$^{\dagger}$ & 1.635$^{\dagger}$ & 78.41$^{\dagger}$  & 61.9$^{\dagger}$ & 18.8$^{\dagger}$ & 1.439$^{\dagger}$ & 56.23$^{\dagger}$ & 54.3$^{\dagger}$ \\
    SAR3D~\cite{chen2025sar3d} & 16.7 & 1.374 & 16.89  & 38.6 & 20.0 & 0.650 & 15.84 & 29.5 \\
    Trellis~\cite{xiang2025structured} & 23.4 & 0.302 & 4.27  & 27.5 & 26.8 & 0.175 & 2.67 & 23.1  \\
    ShapeLLM-Omni~\cite{ye2025shapellm} & 19.8 & 0.451 & 6.73  & 34.1 & 22.7 & 0.249 & 3.27 & 27.7 \\
    \textbf{AR3D-R1} & \textbf{28.5} & \textbf{0.194} & \textbf{2.74} & \textbf{25.9} &  \textbf{29.3} & \textbf{0.156} & \textbf{1.85}  & \textbf{20.4}  \\
    \bottomrule
    \end{tabular}
    }
\end{table*}

\textbf{Hi-GRPO.}
As shown in Figure~\ref{pipeline}, our RL paradigm decomposes each training iteration into two steps, progressing from coarse geometry to fine-grained appearance. 
In the first step, the model first performs semantic planning at the global geometry level, guided by the 3D prompt and high-level instruction. This semantic reasoning serves three purposes: (1) understanding object subcategories to better capture generation intent, (2) establishing spatial layouts of key components to prevent geometric deviations, and (3) elaborating ambiguous terms to improve generation quality. This process produces $|s_i|$ semantic tokens $\{s_{i,1}, s_{i,2}, \ldots, s_{i,|s_i|}\}$, where $i \in \{0, \ldots, G-1\}$.
In Figure~\ref{pipeline}, the semantic reasoning determines the spatial layout and proportions of components (petals, stamen, stem), ensuring ``the proportion of the flower is balanced'', and specifies color distribution as ``a gradient of pink from the center to the outer''.
The 3D prompt, semantic reasoning, and mesh start token \texttt{<mesh\_start>} are then fed into the model to generate 3D tokens grid by grid, where each grid depends on previous ones. This yields $M$ 3D tokens $\{t_{i,1}, t_{i,2}, \ldots, t_{i,M}\}$, where $M$ denotes the compressed grid count. The coarse 3D shape is obtained via decoding and reconstruction. 
As shown in Figure~\ref{pipeline}, this produces geometrically consistent flower shape with accurate colors.
In the second step, conditioned on the prompt, semantic reasoning, and a low-level visual instruction, the model generates visual reasoning that focuses on refining local appearance through: (1) detailed component textures and interactions, and (2) local attributes such as element counts and symmetry. This produces $|v_i|$ visual tokens $\{v_{i,1}, v_{i,2}, \ldots, v_{i,|v_i|}\}$. As shown in Figure~\ref{pipeline}, this step clarifies the petal textures, stamen-petal spatial relations, and leaf counts. Finally, The model then generates the refined 3D object tokens $\{o_{i,1}, o_{i,2}, \ldots, o_{i,M}\}$.
We adopt the loss formulation from Sec.~\ref{sec:rl_agl} with two modifications: (1) the reward from step 2 is backpropagated to the step 1 via $R_{\text{high}} = R_{\text{high}} + \lambda \cdot R_{\text{low}}$, allowing final quality to supervise global planning through configurable weight $\lambda$; (2) each step independently computes advantages and policy losses from its own rewards, yielding total loss $L = L_{\text{high}} + L_{\text{low}}$.

\textbf{Reward Ensemble Design.}
Text-to-3D quality assessment requires multi-view evaluation across multiple dimensions: aesthetic quality, prompt alignment, component completeness, and appearance consistency. To this end, we introduce tailored reward ensembles for Hi-GRPO. Multiple reward functions across steps effectively prevent reward hacking. As shown in Figure~\ref{reward_ensemble}, our ensemble includes the following expert models, with some shared between steps:

\textbullet\ \textbf{Human Preference Model.}
We adopt HPS V2.1~\cite{wu2023human} mentioned in Sec.~\ref{sec:reward} and apply it to both steps, computing the maximum prompt-view similarity across all viewpoints for the coarse shape and refined object. This yields human preference scores $R_1^{\text{HPM}}$ and $R_2^{\text{HPM}}$ as global reward signals.

\textbullet\ \textbf{Unified Reward Model.}
Given that HPS relies on similarity computation at 224×224 resolution, we introduce UnifiedReward~\cite{wang2025unified} to evaluate prompt relevance and aesthetic quality. In step 1, UnifiedReward Think-qwen-7B~\cite{wang2025unified} scores prompt-coarse shape alignment (1-5) across views, taking the maximum as $R_1^{\text{unified}}$ for geometric supervision. In step 2, we further evaluate the appearance quality. UnifiedReward-2.0-qwen-7b scores textured objects across views on logical coherence, style appeal, and prompt alignment (1-5 each). The maximum sum yields $R_2^{\text{unified}}$.

\textbullet\ \textbf{2D Large Multi-modal Model.}
Since existing specialized reward models inadequately handle 3D consistency verification, we adopt Qwen2.5-VL~\cite{bai2025qwen2} for its strong multi-view 3D understanding. Following the coarse-to-fine progression, in step 1, Qwen2.5-VL-7B verifies whether the generated shape matches the object category, assigning a binary score $R_1^{\text{consist}}$ (0/1) for geometric constraint. In step 2, Qwen2.5-VL-7B assesses appearance consistency across views: color smoothness, material realism and coherence, and texture rationality, each scored 0-1 with sum $R_2^{\text{consist}}$.

\textbullet\ \textbf{3D Large Multi-modal Model.}
However, 2D LMMs struggle to accurately detect 3D components from multi-view observations and may misidentify parts. To address this, in step 2 we sample the refined mesh into a 3D point cloud and employ ShapeLLM~\cite{qi2024shapellm} to directly detect the existence (binary 0/1) and completeness (scored 0-1) of key components mentioned in the prompt, summing to $R_2^{\text{part}}$.

\textbf{Experimental Analysis and Insights.} 
We develop AR3D-R1 via Hi-GRPO training, the first RL-enhanced 3D autoregressive generation model, and evaluate it against existing text-to-3D methods on MME-3DR and sampled Toys4K test set. Table~\ref{tab:performance} shows quantitative results, and Figure~\ref{compare} and~\ref{diffstage} presents qualitative comparisons. For reward evaluation, we sample six views per object. Key findings:

\textbullet\ \textit{Hi-GRPO effectively enables hierarchical reasoning from global to local in 3D autoregressive generation. AR3D-R1 exhibits a coarse-to-fine progression, evolving from rough shapes to refined 3D objects.} 
As illustrated in Figure~\ref{diffstage}, the model first generates basic shapes of the truck, helmet, and computer, then refines local details including colors, textures, and part structures to match the prompt.

\textbullet\ \textit{AR3D-R1 achieves superior performance on both MME-3DR and sampled Toys4K test set.}
As shown in Table~\ref{tab:performance}, AR3D-R1 attains optimal results across benchmarks and demonstrates improved cross-dataset robustness compared to other text-to-3D models. 
As illustrated in Figure~\ref{compare}, AR3D-R1 produces high-quality meshes across diverse categories, including animals and mechanical objects.

\section{Conclusion}
This paper presents the first systematic study of reinforcement learning for text-to-3D autoregressive generation. We identify key factors in reward design, RL algorithms, text-to-3D benchmarks and RL paradigms. Our proposed Hi-GRPO leverages the hierarchical nature of 3D generation through dedicated reward ensembles, optimizing global-to-local generation from coarse shapes to fine textures. Building on insights, we develop AR3D-R1, the first RL-enhanced text-to-3D model, which demonstrates superior performance on our proposed MME-3DR and existing benchmarks, achieving significant improvements in geometry consistency and texture quality. Our work provides valuable insights for research in RL-driven 3D generation.

\section*{Contributions of Co-first Authors}
\label{F}
\begin{itemize}
    \item Yiwen Tang: Conducted experiments, primary manuscript writing, and idea discussion.
    \item Zoey Guo: Contributed to experiments and manuscript refinement.
    \item Kaixin Zhu: Contributed to experiments and manuscript refinement.
    \item Ray Zhang: Project leader, responsible for idea proposal, experimental design, and overall manuscript planning.
\end{itemize}

\clearpage
\newpage
\appendix
\section*{Overview}
\begin{itemize}
    \item \cref{C}: Experimental settings.
    \item \cref{D}: Details of Hi-GRPO.
    \item \cref{E}: Ablation study.
    \item \cref{G}: Additional visualizations.
\end{itemize}

\begin{table*}[tb!]
\centering
\caption{\textbf{Quantitative comparisons using Toys4k for Reward Analysis.}}
\label{tab:reward_ablation}
\scalebox{0.9}{
\begin{tabular}{cccccccc@{\hspace{1em}}c}
\toprule
\multicolumn{3}{c}{Step 1 Reward} & \multicolumn{4}{c}{Step 2 Reward} & \multicolumn{2}{c}{Metrics} \\
\cmidrule(r){1-3} \cmidrule(l){4-7} \cmidrule(l){8-9}
$R_1^{\text{HPM}}$ &  $R_1^{\text{unified}}$ & $R_1^{\text{consist}}$ & $R_2^{\text{HPM}}$ &  $R_2^{\text{unified}}$ & $R_2^{\text{consist}}$ & $R_2^{\text{part}}$ & CLIP Score$\uparrow$ & KD$_{\text{incep}}$$\downarrow$ \\
\midrule
- & - & - & \checkmark & \checkmark & - & - & 25.0 & 0.235 \\
- & - & - & \checkmark & \checkmark & \checkmark & - & 25.7 & 0.223 \\
\checkmark & \checkmark & - & \checkmark & \checkmark & \checkmark & - & 27.8 & 0.194 \\
\checkmark & \checkmark & \checkmark & \checkmark & \checkmark & \checkmark & - & 28.3 & 0.182 \\
\checkmark & \checkmark & - & \checkmark & \checkmark & \checkmark & \checkmark & 28.6 & 0.178 \\
\checkmark & \checkmark & \checkmark & \checkmark & \checkmark & \checkmark & \checkmark & \textbf{29.3} & \textbf{0.156} \\
\bottomrule
\end{tabular}
}
\end{table*}

\section{Experimental Settings}
\label{C}
We employ ShapeLLM-Omni~\cite{ye2025shapellm} as the base model with a learning rate of $1\times10^{-6}$, $\beta$ of $0.01$, and group size of 8. Training is conducted on 8 GPUs with a batch size of 1 per device and gradient accumulation over 2 steps. The model is trained for 1{,}200 steps. The configurable weight $\lambda$ for supervising global planning with final quality is set to $1.0$.
Our reward models are deployed via the vLLM API framework.
We select training prompt from Objaverse-XL~\cite{deitke2023objaverse}, HSSD~\cite{khanna2024habitat}, and ABO~\cite{collins2022abo}, and evaluate our method on Toys4K~\cite{stojanov2021using}.

\textbullet\ \textbf{Objaverse-XL:} Objaverse-XL is one of the largest 3D object datasets currently available, comprising over 10 million 3D objects sourced from diverse platforms including GitHub, Thingiverse, Sketchfab and Polycam. The dataset undergoes rigorous deduplication and rendering validation, covering a range of categories and fine-grained attributes. 

\textbullet\ \textbf{HSSD:} HSSD contains 211 high-quality indoor synthetic 3D scenes with approximately 18,656 real-world object models, emphasizing indoor layouts, semantic structures, and object relationships.

\textbullet\ \textbf{ABO:} ABO focuses on real-world household objects and provide approximately 147,000 product listings, nearly 400,000 catalog images, and about 8,000 3D models with rich material, geometric, and attribute annotations.

\textbullet\ \textbf{Toys4K:} Toys4K includes approximately 4,000 3D object instances spanning around 105 categories, featuring diverse categories and significant shape variations.

\section{Details of Hi-GRPO}
\label{D}

\subsection{Two-Step Generation Process}

\textbf{Step 1}: The model generates semantic reasoning tokens $\mathbf{s}_i = \{s_{i,1}, \ldots, s_{i,|s_i|}\}$ for global geometric planning, followed by 3d tokens $\mathbf{t}_i = \{t_{i,1}, \ldots, t_{i,M}\}$, where $M$ is the number of compressed grids. The coarse triangular mesh $\mathcal{M}_i^{(1)}$ is decoded through the VQVAE decoder.

\textbf{Step 2}: Conditioned on semantic reasoning, the model generates visual reasoning tokens $\mathbf{v}_i = \{v_{i,1}, \ldots, v_{i,|v_i|}\}$ focused on local details, followed by 3d tokens $\mathbf{o}_i = \{o_{i,1}, \ldots, o_{i,M}\}$, which are decoded into mesh $\mathcal{M}_i^{(2)}$.

\subsection{Hierarchical Reward Ensemble Design}

\subsubsection{Human Preference Model}

We adopt HPS V2.1~\cite{wu2023human} in both steps. For generated 3D objects rendered from 6 uniformly distributed viewpoints $\{\mathbf{v}_1, \ldots, \mathbf{v}_6\}$, we compute text-image similarity at each viewpoint and take the maximum:
\begin{equation}
R_i^{\text{HPM},k} = \max_{j=1,\ldots,6} \text{HPS}(\mathbf{x}, \text{Render}(\mathcal{M}_i^{(k)}, \mathbf{v}_j))
\end{equation}
This reward evaluates human preference with range [0, 1].

\subsubsection{Unified Reward Model}

\textbf{Step 1}: UnifiedReward Think-qwen-7B~\cite{wang2025unified} evaluates geometric alignment between prompts and coarse shapes. Each of the 6 viewpoints is scored (1-5), and the maximum is:
\begin{equation}
R_i^{\text{unified},1} = \max_{j=1,\ldots,6} f_{\text{UR-Think}}(\mathbf{x}, \text{Render}(\mathcal{M}_i^{(1)}, \mathbf{v}_j))
\end{equation}
This reward evaluates prompt alignment with range [1, 5].

\textbf{Step 2}: UnifiedReward-2.0-qwen-7b~\cite{wang2025unified} performs three-dimensional evaluation of textured objects: (1) prompt alignment (1-5), (2) logical coherence (1-5), (3) style appeal (1-5). The maximum sum across 6 viewpoints:
\begin{equation}
\begin{aligned}
R_i^{\text{unified},2} 
= \max_{j=1,\ldots,6}
    \sum_{\ell \in A_\text{app}}
    f_{\text{UR}}^{(\ell)}(
        \mathbf{x}, \text{Render}(\mathcal{M}_i^{(2)}, \mathbf{v}_j)
    )
\end{aligned}
\end{equation}
$\mathcal{A}_{\text{app}} = \{\text{logic},\, \text{style},\, \text{align}\}.$
This reward evaluates 3 dimensions with range [3, 15].

\subsubsection{2D Large Multi-modal Model}

\textbf{Step 1}: Qwen2.5-VL-7B~\cite{bai2025qwen2} verifies whether the generated shape matches the object category in the prompt based on joint observation of 6 viewpoints:
\begin{equation}
R_i^{\text{consist},1} = f_{\text{Qwen}}^{\text{category}}(\mathbf{x}, \{\text{Render}(\mathcal{M}_i^{(1)}, \mathbf{v}_j)\}_{j=1}^6)
\end{equation}
This reward evaluates category matching with range $\{0, 1\}$.

\textbf{Step 2}: Qwen2.5-VL-7B~\cite{bai2025qwen2} evaluates three dimensions of cross-view appearance consistency: (1) color smoothness (0-1), (2) material realism and coherence (0-1), (3) texture rationality (0-1):
\begin{equation}
R_i^{\text{consist},2}
= \sum_{\ell \in \mathcal{A}_{\text{app}}}
f_{\text{Qwen}}^{(\ell)}\!\bigl(
  \mathbf{x},
  \{\text{Render}(\mathcal{M}_i^{(2)}, \mathbf{v}_j)\}_{j=1}^6
\bigr).
\end{equation}
$\mathcal{A}_{\text{app}} = \{\text{color},\, \text{material},\, \text{texture}\}.$
This reward evaluates 3 dimensions with range [0, 3].

\begin{table*}[tb!]
\centering
\caption{\textbf{Quantitative comparisons using Toys4k for Different RL Paradigms.}}
\label{tab:grpo_ablation}
\scalebox{0.9}{
\begin{tabular}{cccccc@{\hspace{1em}}c}
\toprule
\multicolumn{5}{c}{Training Strategy} & \multicolumn{2}{c}{Metrics} \\
\cmidrule(r){1-5} \cmidrule(l){6-7}
GRPO & Textual Reasoning & Step1 Reward & Step2 Reward & Hi-GRPO & CLIP Score$\uparrow$ & KD$_{\text{incep}}$$\downarrow$ \\
\midrule
- & - & - & - & -  & 22.7 & 0.249 \\
\checkmark & - & - & - & - & 24.3 & 0.237 \\
\checkmark & \checkmark & - & - & - & 25.2 & 0.228 \\
\checkmark & \checkmark & \checkmark & -  & - & 24.8 & 0.235 \\
\checkmark & \checkmark & - &  \checkmark & - & 26.0 & 0.214 \\
- & - & - & -  & \checkmark & \textbf{28.7} & \textbf{0.182} \\
\bottomrule
\end{tabular}
}
\end{table*}

\subsubsection{3D Large Multi-modal Model}
2D LMMs struggle to accurately detect 3D components from multi-view observations. To obtain accurate component completeness assessment, we employ direct evaluation based on 3D point clouds in step 2.

\textbf{1) Mesh to Dense Point Cloud Sampling}: The refined triangular mesh $\mathcal{M}_i^{(2)} = (\mathcal{V}^{(2)}, \mathcal{F}^{(2)}, \mathcal{T})$ is converted to dense point cloud $\mathcal{P}_i$. The sampling process:

\begin{enumerate}
\item \textbf{Area-Weighted Sampling}: For each triangle face $f \in \mathcal{F}^{(2)}$, allocate sample points $n_f = \lceil \rho \cdot A_f \rceil$ based on area $A_f$, where $\rho$ is the sampling density parameter
\item \textbf{Barycentric Uniform Sampling}: Within face $f = (v_1, v_2, v_3)$, generate random barycentric coordinates $(\alpha, \beta, \gamma)$ satisfying $\alpha + \beta + \gamma = 1$ and $\alpha, \beta, \gamma \geq 0$. Sample point coordinates: $\mathbf{p} = \alpha \mathbf{v}_1 + \beta \mathbf{v}_2 + \gamma \mathbf{v}_3$
\item \textbf{Texture Color Sampling}: Interpolate UV coordinates using barycentric coordinates $\mathbf{uv} = \alpha \mathbf{uv}_1 + \beta \mathbf{uv}_2 + \gamma \mathbf{uv}_3$, and sample RGB color from texture map $\mathcal{T}$
\end{enumerate}
The result is point cloud $\mathcal{P}_i = \{(\mathbf{p}_k, \mathbf{c}_k)\}_{k=1}^{N_p}$, where $\mathbf{p}_k \in \mathbb{R}^3$ is position, $\mathbf{c}_k \in \mathbb{R}^3$ is RGB color.

\textbf{2) Per-Component Evaluation}:
Parse prompt $\mathbf{x}$ to extract component list $\mathcal{C} = \{c_1, \ldots, c_{N_c}\}$ and expected quantities $\{n_1, \ldots, n_{N_c}\}$.
ShapeLLM~\cite{qi2024shapellm} processes point cloud $\mathcal{P}_i$ and component queries. For each component $c_p$:
\begin{itemize}
\item Existence: $e_p \in \{0, 1\}$ determines the existence.
\item Completeness: $q_p \in [0, 1]$ evaluates geometric completeness, shape correctness, and quantity matching
\end{itemize}
Average scores across $N_c$ components:
\begin{equation}
R_i^{\text{part},2} = \frac{1}{N_c} \sum_{p=1}^{N_c} (e_p + q_p)
\end{equation}

This reward evaluates 2 dimensions per component (existence + completeness), with averaged range [0, 2].

\subsubsection{Dimension-Normalized Reward Ensemble}

Each reward is normalized by its number of evaluation dimensions to ensure fair contribution:

\textbf{Step 1 Total Reward}:
\begin{equation}
R_i^{\text{high}} = R_i^{\text{HPM},1} + R_i^{\text{unified},1} + R_i^{\text{consist},1}
\end{equation}

\textbf{Step 2 Total Reward}:
\begin{equation}
R_i^{\text{low}} = R_i^{\text{HPM},2} + \frac{R_i^{\text{unified},2}}{3} + \frac{R_i^{\text{consist},2}}{3} + \frac{R_i^{\text{part},2}}{2}
\end{equation}

This normalization strategy ensures: (1) each reward's contribution is proportional to its number of evaluation dimensions; (2) multi-dimensional evaluations do not dominate through simple summation; (3) the system remains stable when adding or removing rewards.
Rewards from step 2 are backpropagated to step 1 through weight $\lambda$:
\begin{equation}
\tilde{R}_i^{\text{high}} = R_i^{\text{high}} + \lambda \cdot R_i^{\text{low}}
\end{equation}
When $\lambda = 1.0$, the high-level step is directly supervised by final output quality.
For each step, advantages are normalized within prompt groups to eliminate reward scale differences across prompts:
\begin{equation}
A_i^{(1)} = \frac{\tilde{R}_i^{\text{high}} - \mu_g^{(1)}}{\sigma_g^{(1)} + \epsilon}, \quad A_i^{(2)} = \frac{R_i^{\text{low}} - \mu_g^{(2)}}{\sigma_g^{(2)} + \epsilon}
\end{equation}
where $\mu_g^{(k)} = \frac{1}{G}\sum_{j=1}^{G} R_j^{(k)}$, $\sigma_g^{(k)} = \sqrt{\frac{1}{G}\sum_{j=1}^{G} (R_j^{(k)} - \mu_g^{(k)})^2}$, $\epsilon = 10^{-4}$.

\begin{figure*}[tb!]
  \centering
  \includegraphics[width=\linewidth]{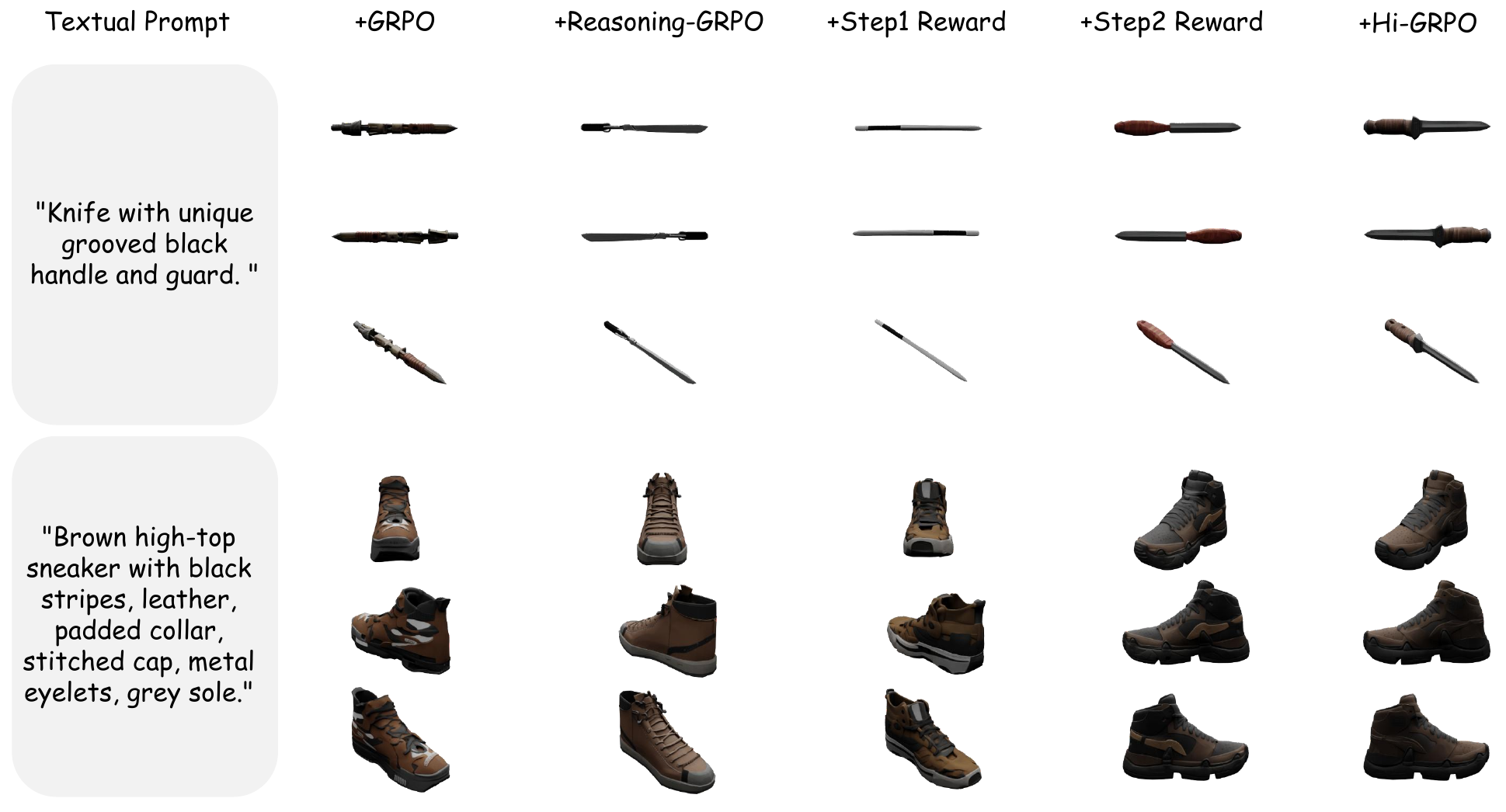}
  \caption{\textbf{Visualization Results of Small Objects for Different RL Paradigms.}}
\label{diffreward_1}
\end{figure*}

\subsection{Loss Computation}

For each step, we compute token-level log probabilities by concatenating the log probabilities of reasoning tokens and mesh tokens. In step 1, the complete sequence log probability concatenates semantic reasoning and coarse mesh generation: $\log \pi_\theta(\mathbf{y}_i^{(1)}) = \text{concat}(\log \pi_\theta(\mathbf{s}_i \mid \mathbf{x}), \log \pi_\theta(\mathbf{t}_i \mid \mathbf{x}, \mathbf{s}_i))$. In step 2, it concatenates visual reasoning and refined mesh generation: $\log \pi_\theta(\mathbf{y}_i^{(2)}) = \text{concat}(\{\log \pi_\theta(v_{i,t} \mid \mathbf{x}, \mathbf{s}_i, v_{i,<t})\}_{t=1}^{|\mathbf{v}_i|}, \{\log \pi_\theta(o_{i,t} \mid \mathbf{x}, \mathbf{v}_i, o_{i,<t})\}_{t=1}^{M})$. Reference policy log probabilities $\log \pi_{\text{ref}}(\mathbf{y}_i^{(k)})$ are computed similarly.

For stage $k \in \{1,2\}$, the complete loss function is:

\begin{equation}
\begin{aligned}
\mathcal{L}^{(k)} = -\mathbb{E}_{q\sim\mathcal{D}, \{y_i^{(k)}\}_{i=1}^{G}\sim\pi_{\theta_{\text{old}}}}\Bigg[&\frac{1}{\sum_{i=1}^{G}T_i^{(k)}}\sum_{i=1}^{G}\sum_{t=1}^{T_i^{(k)}}\\ m_{i,t}^{(k)} \Bigg(\min\Big(r_{i,t}^{(k)}(\theta) A_i^{(k)},
\text{clip}(r_{i,t}^{(k)}(\theta), \\1-\varepsilon_{\text{low}}, 1+\varepsilon_{\text{high}}) A_i^{(k)}\Big) - \beta \cdot \text{KL}_{i,t}^{(k)}\Bigg)\Bigg]
\end{aligned}
\end{equation}
We highlight and describe key components as follows:

\textbullet\ \textbf{Policy Ratio}:
\begin{equation}
r_{i,t}^{(k)}(\theta) = \frac{\pi_\theta(y_{i,t}^{(k)} | \mathbf{y}_{i,<t}^{(k)})}{\pi_{\theta_{\text{old}}}(y_{i,t}^{(k)} | \mathbf{y}_{i,<t}^{(k)})}
\end{equation}

\textbullet\ \textbf{Decoupled Clipping}: Asymmetric clipping thresholds $\varepsilon_{\text{low}}$ and $\varepsilon_{\text{high}}$. The higher threshold allows low-probability tokens greater probability increase space, promoting exploration and preventing entropy collapse.

\textbullet\ \textbf{Token-Level Averaging}: The loss is normalized by the token count $\sum_{i=1}^{G}T_i^{(k)}$, where $T_i^{(k)} = \sum_{t=1}^{T_{\max}} m_{i,t}^{(k)}$ is the number of valid tokens and $m_{i,t}^{(k)}$ is the completion mask.

\textbullet\ \textbf{KL Regularization}: Token-level KL divergence
\begin{equation}
\text{KL}_{i,t}^{(k)} = \frac{\pi_{\text{ref}}(y_{i,t}^{(k)} | \mathbf{y}_{i,<t}^{(k)})}{\pi_\theta(y_{i,t}^{(k)} | \mathbf{y}_{i,<t}^{(k)})} - \log\frac{\pi_{\text{ref}}(y_{i,t}^{(k)} | \mathbf{y}_{i,<t}^{(k)})}{\pi_\theta(y_{i,t}^{(k)} | \mathbf{y}_{i,<t}^{(k)})} - 1
\end{equation}
with penalty coefficient $\beta=0.01$ prevents the policy from deviating too far from the reference.
The two steps compute losses independently. The total optimization objective:
\begin{equation}
\mathcal{L}_{\text{total}} = \mathcal{L}^{(1)} + \mathcal{L}^{(2)}
\end{equation}

\begin{figure*}[tb!]
  \centering
  \includegraphics[width=\linewidth]{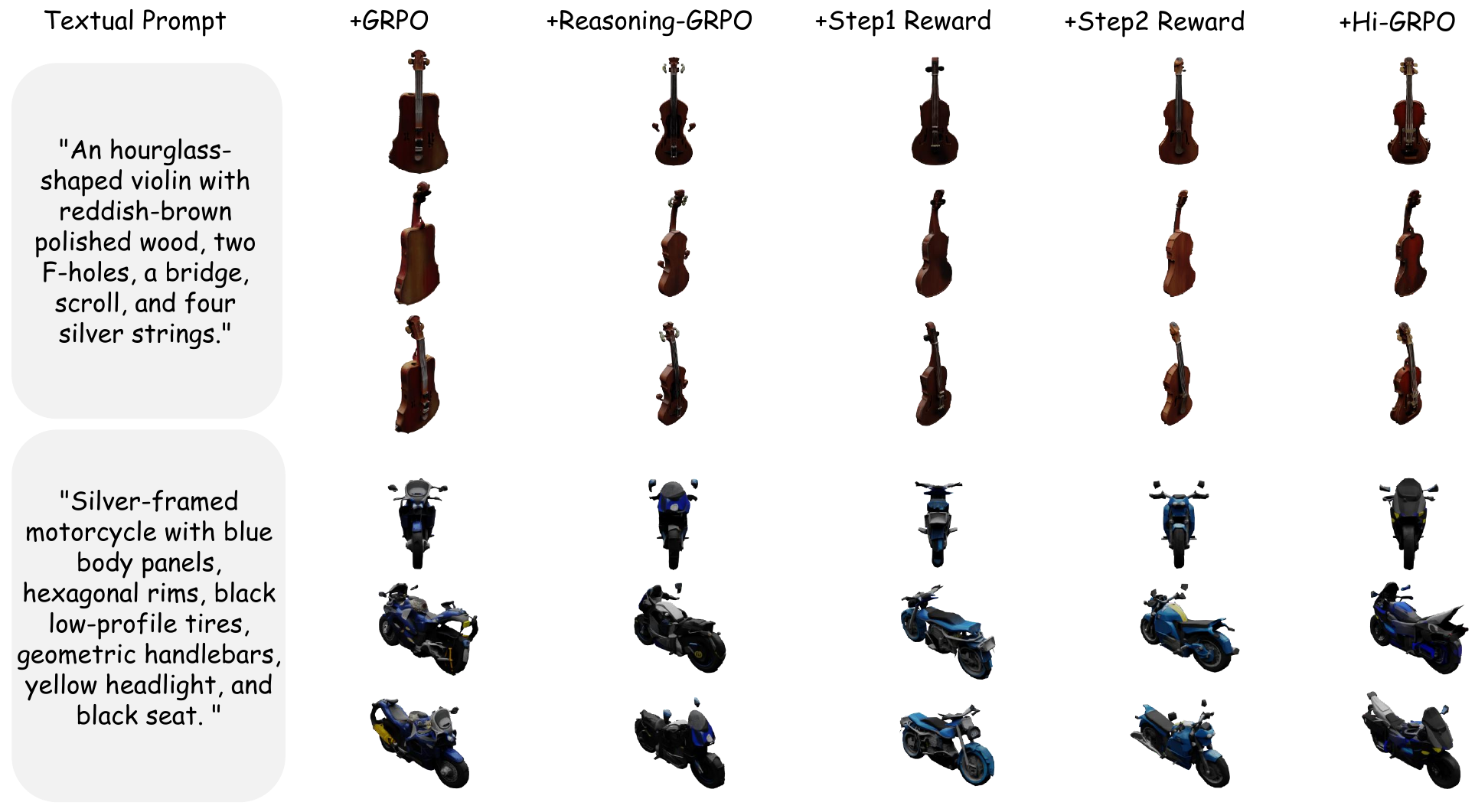}
  \caption{\textbf{Visualization Results of Large Objects for Different RL Paradigms.}}

\label{diffreward_2}
\end{figure*}

\section{Ablation Study}
\label{E}
\subsection{Reward Analysis.}
This section investigates reward function selection and combination for AR3D-R1. Table~\ref{tab:reward_ablation} presents our findings. We first examine whether Step-2 rewards can simultaneously optimize both generation steps. Results show that rewards from refined objects struggle to control both coarse geometry and fine texture effectively. Even with combined rewards ($R_2^{\text{HPM}}$+$R_2^{\text{unified}}$+$R_2^{\text{consist}}$), improvements remain marginal. However, introducing step-specific rewards, adding $R_1^{\text{HPM}}$+$R_1^{\text{unified}}$ for Step 1, yields substantial gains, improving CLIP scores by 2.1 point. Notably, component-level rewards prove critical for ensuring correct part positioning, quantity, and structural plausibility.

\subsection{Effectiveness of Hi-GRPO.}
To validate Hi-GRPO, we conduct ablation studies using the baseline GRPO algorithm. Table~\ref{tab:grpo_ablation} presents quantitative results, while Figures~\ref{diffreward_1} and~\ref{diffreward_2} provide extensive visualizations. We first compare direct 3D token optimization against textual reasoning-guided GRPO, both evaluated with HPSV2.1+UnifiedReward+Qwen2.5-VL reward system. Quantitatively, textual reasoning yields a 0.9-point CLIP score improvement, while qualitatively it enables effective global planning for both large and small objects. We then examine Hi-GRPO's hierarchical reward ensemble by separately applying Step-1 and Step-2 reward systems. Since Step-1 rewards focus on high-level geometric structure, Table~\ref{tab:grpo_ablation} shows performance degradation, with visualizations revealing noticeably reduced texture fidelity. Ultimately, the hierarchical RL paradigm of Hi-GRPO, combining global-to-local generation with step-specific reward ensembles, achieves substantial improvements across geometry, fine-grained textures, and prompt alignment.

\begin{figure*}[tb!]
  \centering
  \includegraphics[width=\linewidth]{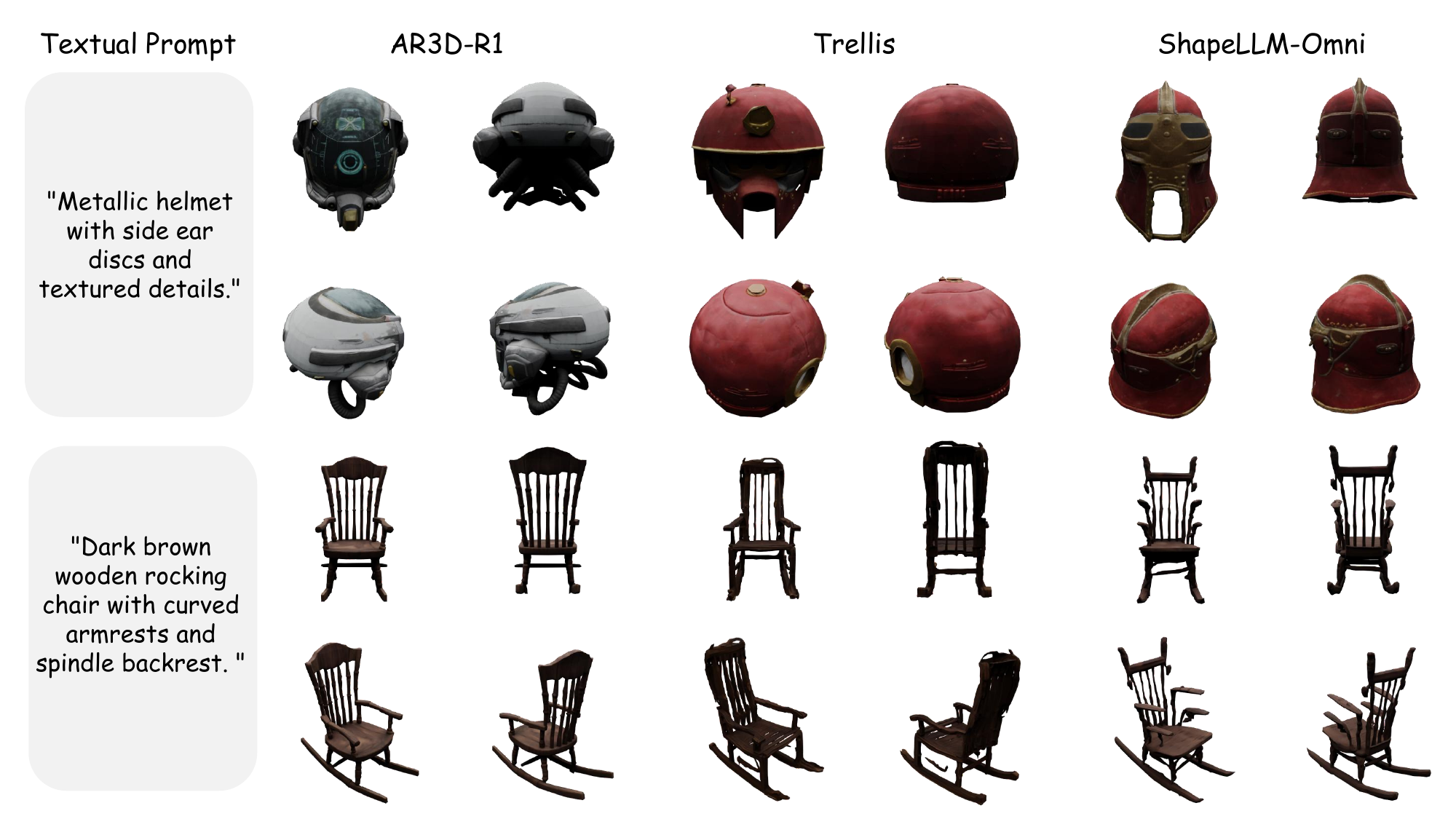}
  \caption{\textbf{Visualization Results of Spatial \&
Structural Geometry in MME-3DR.}}
\label{visual_1}
\end{figure*}

\begin{figure*}[tb!]
  \centering
  \includegraphics[width=\linewidth]{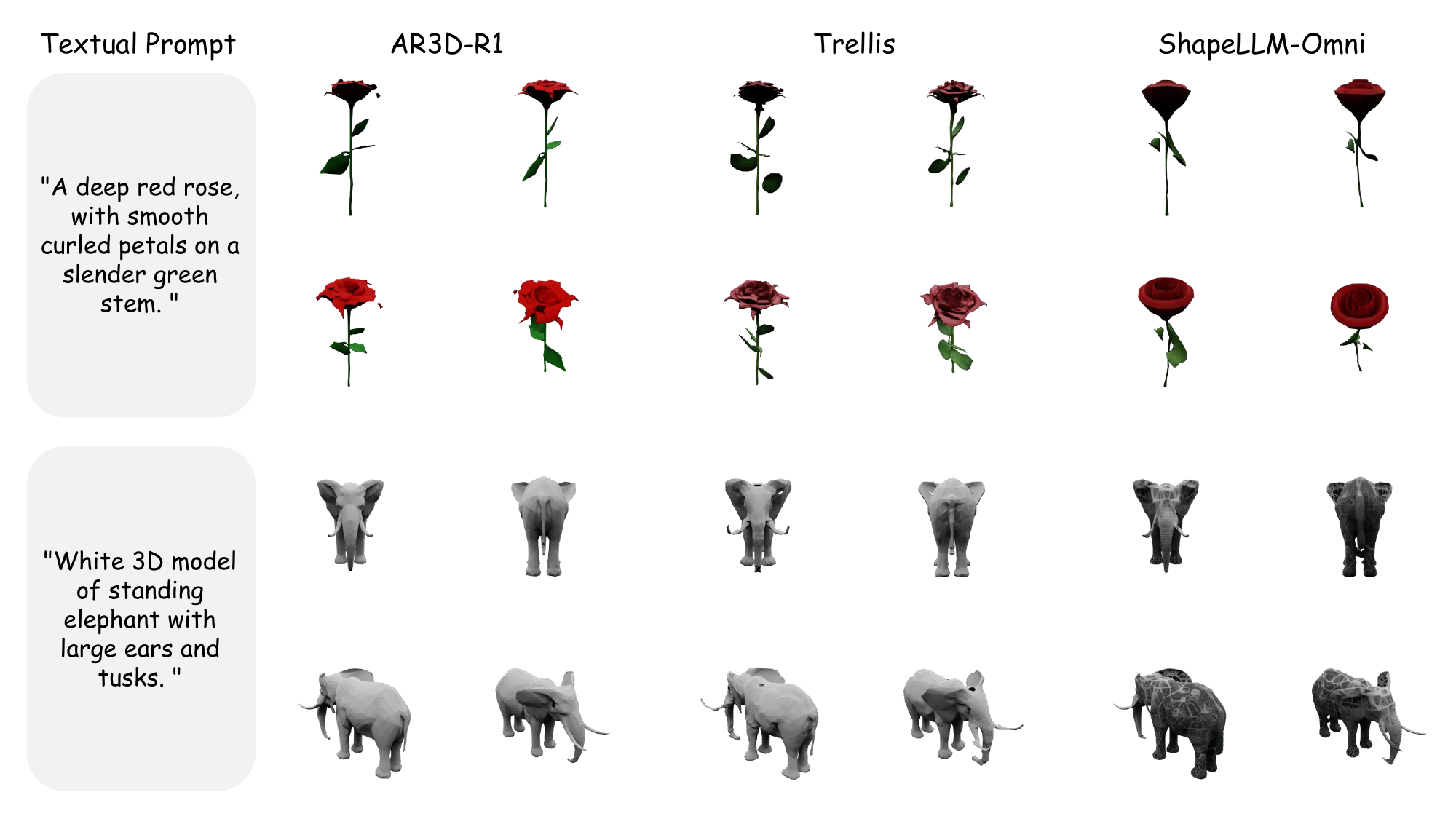}
  \caption{\textbf{Visualization Results of Biological \& Organic Shapes in MME-3DR.}}
\label{visual_2}
\end{figure*}

\begin{figure*}[tb!]
  \centering
  \includegraphics[width=\linewidth]{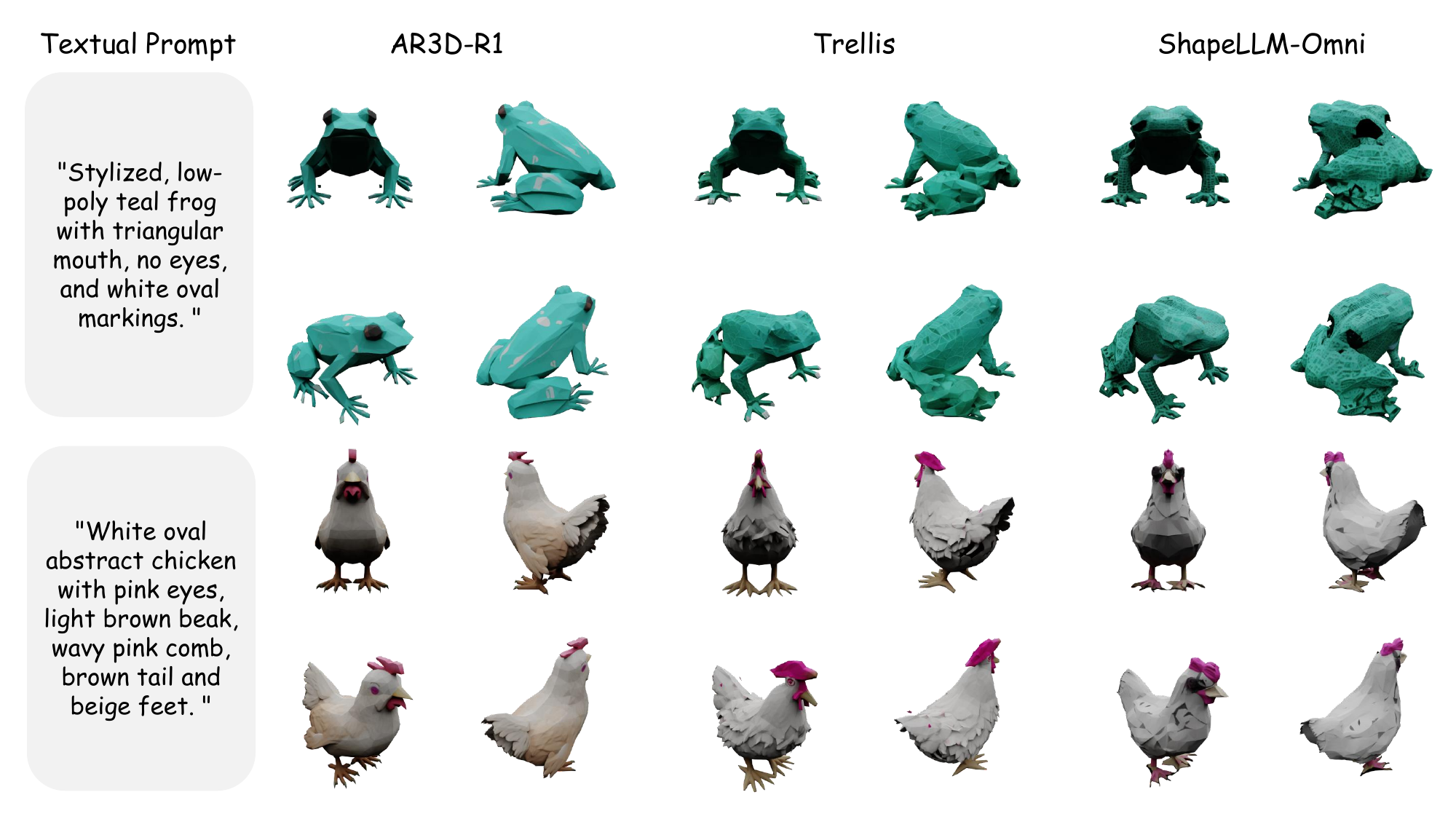}
  \caption{\textbf{Visualization Results of Stylized
Representations in MME-3DR.}}
\label{visual_3}
\end{figure*}

\begin{figure*}[tb!]
  \centering
  \includegraphics[width=\linewidth]{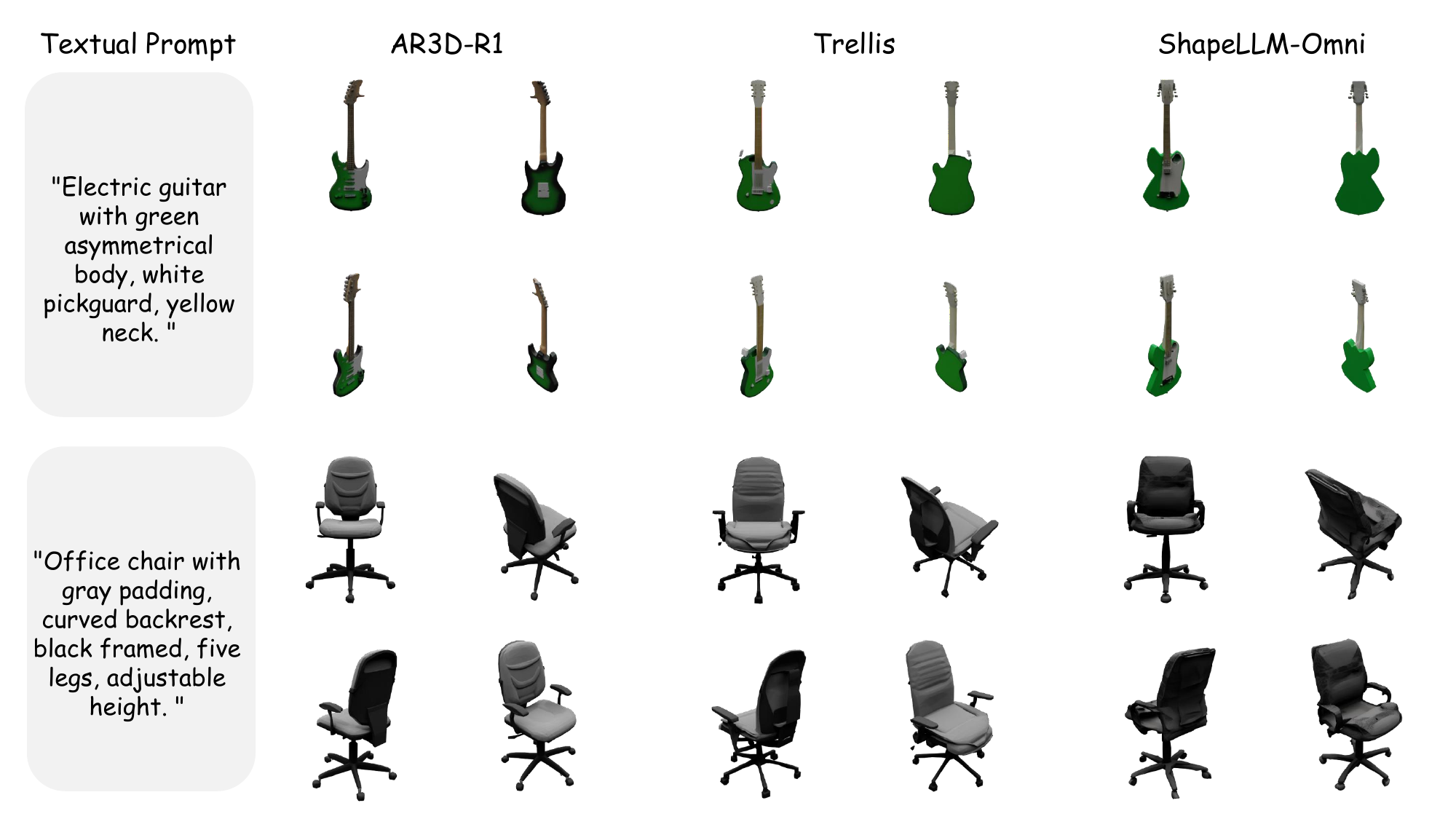}
  \caption{\textbf{Visualization Results of Mechanical Affordances in MME-3DR.}}
\label{visual_4}
\end{figure*}

\begin{figure*}[tb!]
  \centering
  \includegraphics[width=\linewidth]{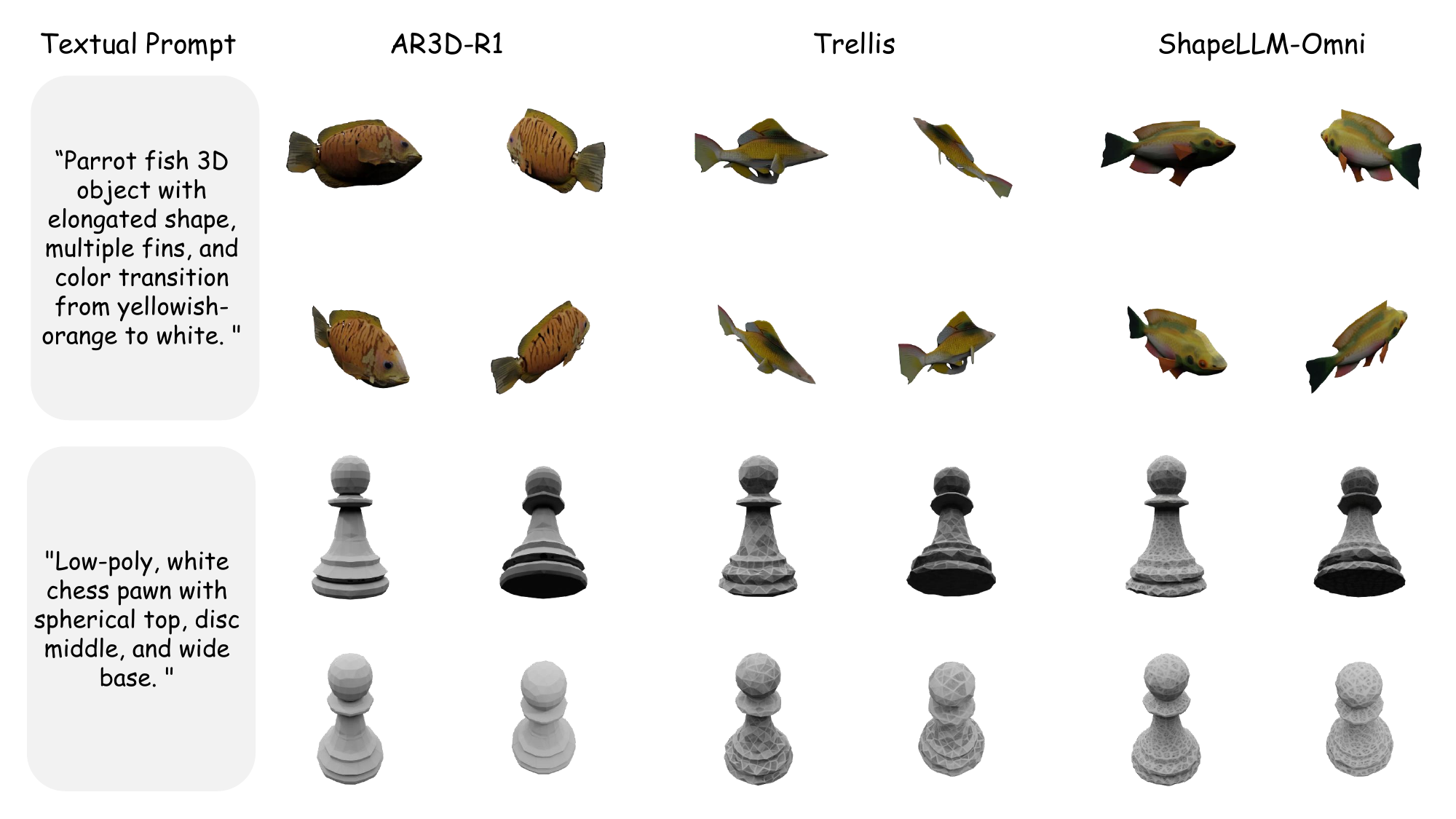}
  \caption{\textbf{Visualization Results of World-Knowledge
Rare Objects in MME-3DR.}}
\label{visual_5}
\end{figure*}

\begin{figure*}[tb!]
  \centering
  \includegraphics[width=\linewidth]{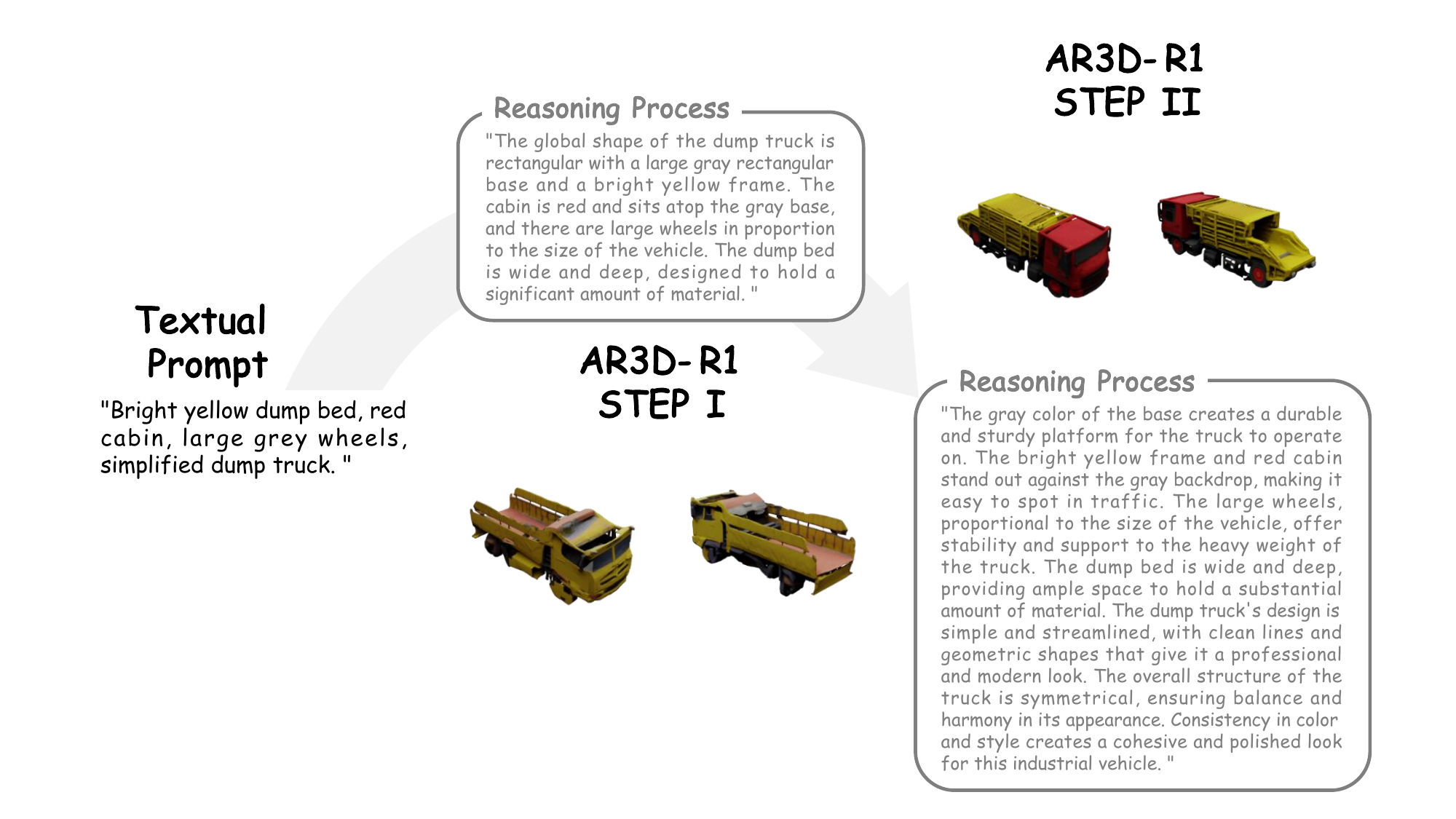}
  
  \caption{\textbf{Visualization of the Two-Step Reasoning Generation Process in Cabin.}}

\label{visual_6}
\end{figure*}

\begin{figure*}[tb!]

  \centering
  \includegraphics[width=\linewidth]{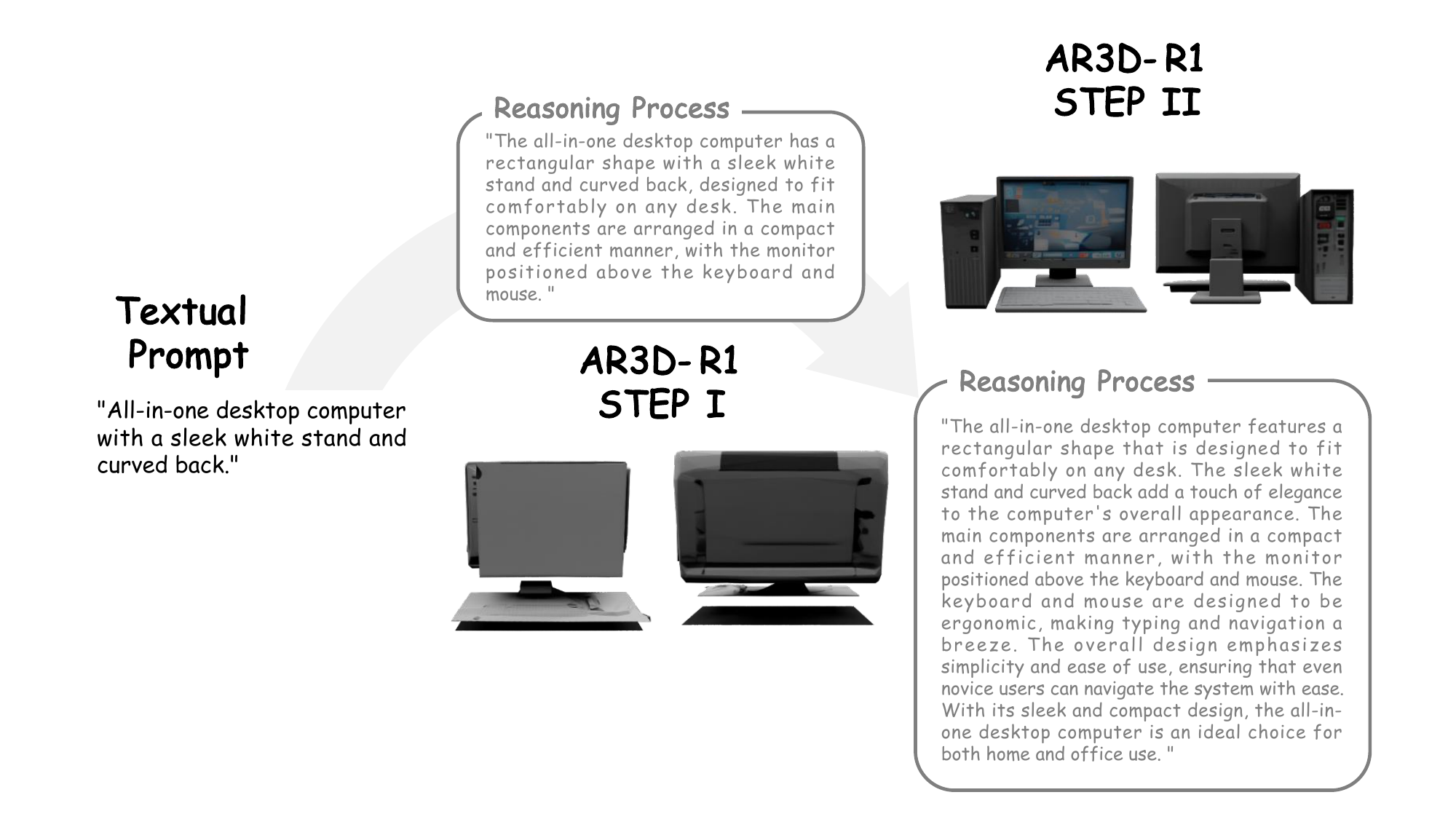}
  
  \caption{\textbf{Visualization of the Two-Step Reasoning Generation Process in Desktop Computer.}}

\label{visual_7}
\end{figure*}

\begin{figure*}[tb!]
  \centering
  \includegraphics[width=\linewidth]{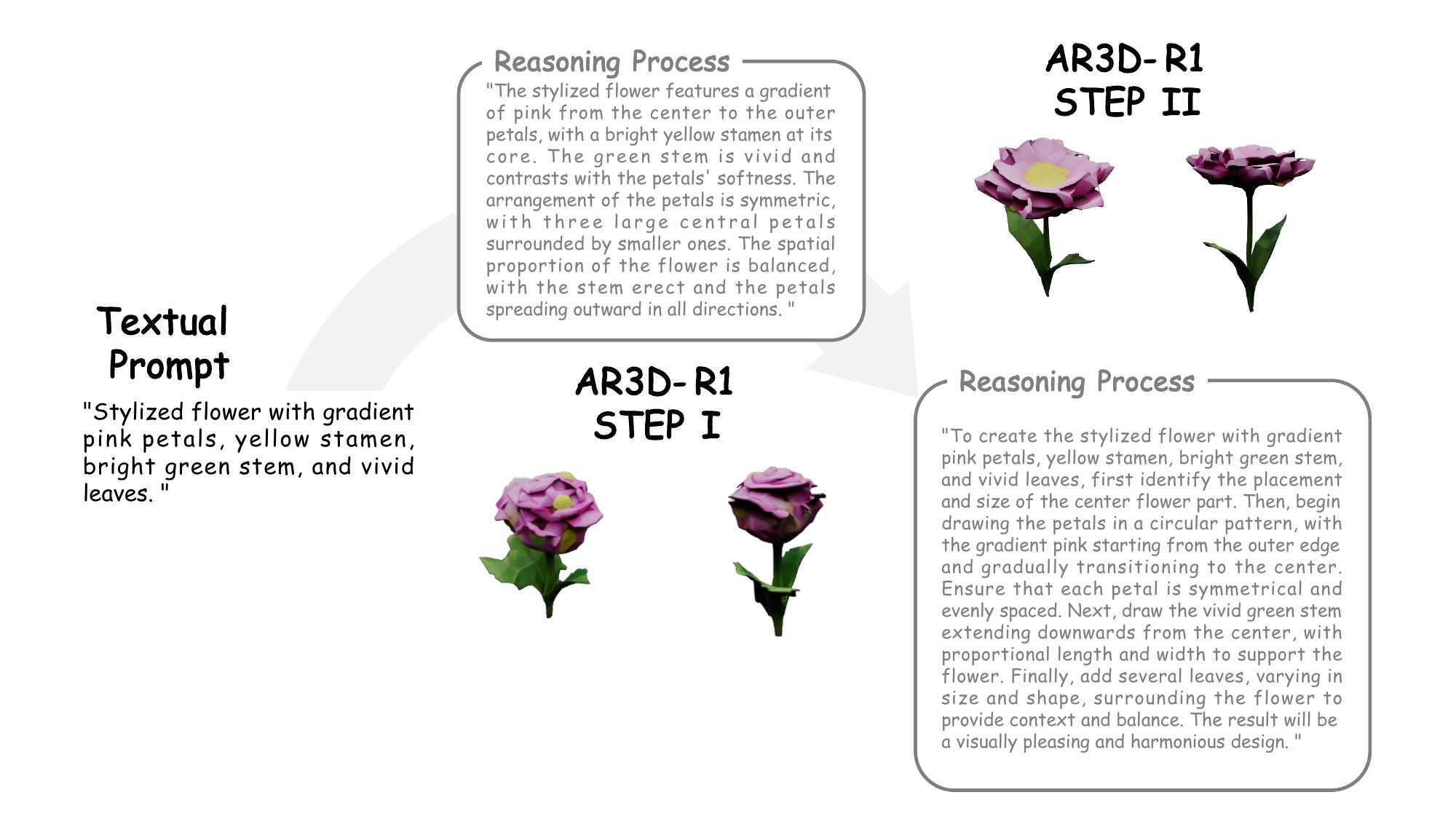}
  
  \caption{\textbf{Visualization of the Two-Step Reasoning Generation Process in Stylized Flower.}}

\label{visual_8}
\end{figure*}

\section{Additional Visualizations}
\label{G}
Figures~\ref{visual_1},~\ref{visual_2},~\ref{visual_3}, ~\ref{visual_4}, and~\ref{visual_5} visualize the generation results of our proposed AR3D-R1, ShapeLLM-Omni, and Trellis across the five categories in our MME-3DR benchmark.
Figures~\ref{visual_6},~\ref{visual_7}, and~\ref{visual_8} visualize AR3D-R1's hierarchical generation process across different object categories.

%% file: main.bbl
\begin{thebibliography}{45}
\providecommand{\natexlab}[1]{#1}
\providecommand{\url}[1]{\texttt{#1}}
\expandafter\ifx\csname urlstyle\endcsname\relax
  \providecommand{\doi}[1]{doi: #1}\else
  \providecommand{\doi}{doi: \begingroup \urlstyle{rm}\Url}\fi

\bibitem[Bai et~al.(2025)Bai, Chen, Liu, Wang, Ge, Song, Dang, Wang, Wang, Tang, et~al.]{bai2025qwen2}
Shuai Bai, Keqin Chen, Xuejing Liu, Jialin Wang, Wenbin Ge, Sibo Song, Kai Dang, Peng Wang, Shijie Wang, Jun Tang, et~al.
\newblock Qwen2. 5-vl technical report.
\newblock \emph{arXiv preprint arXiv:2502.13923}, 2025.

\bibitem[Chen et~al.(2024)Chen, He, Huang, Ye, Chen, Tang, Chen, Cai, Yang, Yu, et~al.]{chen2024meshanything}
Yiwen Chen, Tong He, Di Huang, Weicai Ye, Sijin Chen, Jiaxiang Tang, Xin Chen, Zhongang Cai, Lei Yang, Gang Yu, et~al.
\newblock Meshanything: Artist-created mesh generation with autoregressive transformers.
\newblock \emph{arXiv preprint arXiv:2406.10163}, 2024.

\bibitem[Chen et~al.(2025{\natexlab{a}})Chen, Lan, Zhou, Wang, and Pan]{chen2025sar3d}
Yongwei Chen, Yushi Lan, Shangchen Zhou, Tengfei Wang, and Xingang Pan.
\newblock Sar3d: Autoregressive 3d object generation and understanding via multi-scale 3d vqvae.
\newblock In \emph{Proceedings of the Computer Vision and Pattern Recognition Conference}, pages 28371--28382, 2025{\natexlab{a}}.

\bibitem[Chen et~al.(2025{\natexlab{b}})Chen, Wang, Luo, Wang, Chen, Zhu, Zhang, and Lin]{chen2025meshanything}
Yiwen Chen, Yikai Wang, Yihao Luo, Zhengyi Wang, Zilong Chen, Jun Zhu, Chi Zhang, and Guosheng Lin.
\newblock Meshanything v2: Artist-created mesh generation with adjacent mesh tokenization.
\newblock In \emph{Proceedings of the IEEE/CVF International Conference on Computer Vision}, pages 13922--13931, 2025{\natexlab{b}}.

\bibitem[Chen et~al.(2025{\natexlab{c}})Chen, Tang, Dong, Cao, Hong, Lan, Wang, Xie, Wu, Saito, et~al.]{chen20253dtopia}
Zhaoxi Chen, Jiaxiang Tang, Yuhao Dong, Ziang Cao, Fangzhou Hong, Yushi Lan, Tengfei Wang, Haozhe Xie, Tong Wu, Shunsuke Saito, et~al.
\newblock 3dtopia-xl: Scaling high-quality 3d asset generation via primitive diffusion.
\newblock In \emph{Proceedings of the Computer Vision and Pattern Recognition Conference}, pages 26576--26586, 2025{\natexlab{c}}.

\bibitem[Collins et~al.(2022)Collins, Goel, Deng, Luthra, Xu, Gundogdu, Zhang, Vicente, Dideriksen, Arora, et~al.]{collins2022abo}
Jasmine Collins, Shubham Goel, Kenan Deng, Achleshwar Luthra, Leon Xu, Erhan Gundogdu, Xi Zhang, Tomas F~Yago Vicente, Thomas Dideriksen, Himanshu Arora, et~al.
\newblock Abo: Dataset and benchmarks for real-world 3d object understanding.
\newblock In \emph{Proceedings of the IEEE/CVF conference on computer vision and pattern recognition}, pages 21126--21136, 2022.

\bibitem[Deitke et~al.(2023)Deitke, Liu, Wallingford, Ngo, Michel, Kusupati, Fan, Laforte, Voleti, Gadre, et~al.]{deitke2023objaverse}
Matt Deitke, Ruoshi Liu, Matthew Wallingford, Huong Ngo, Oscar Michel, Aditya Kusupati, Alan Fan, Christian Laforte, Vikram Voleti, Samir~Yitzhak Gadre, et~al.
\newblock Objaverse-xl: A universe of 10m+ 3d objects.
\newblock \emph{Advances in Neural Information Processing Systems}, 36:\penalty0 35799--35813, 2023.

\bibitem[Feng et~al.(2025)Feng, Gong, Li, Guo, Wang, Peng, Wu, Zhang, Wang, and Yue]{feng2025video}
Kaituo Feng, Kaixiong Gong, Bohao Li, Zonghao Guo, Yibing Wang, Tianshuo Peng, Junfei Wu, Xiaoying Zhang, Benyou Wang, and Xiangyu Yue.
\newblock Video-r1: Reinforcing video reasoning in mllms.
\newblock \emph{arXiv preprint arXiv:2503.21776}, 2025.

\bibitem[Guo et~al.(2025{\natexlab{a}})Guo, Yang, Zhang, Song, Zhang, Xu, Zhu, Ma, Wang, Bi, et~al.]{guo2025deepseek}
Daya Guo, Dejian Yang, Haowei Zhang, Junxiao Song, Ruoyu Zhang, Runxin Xu, Qihao Zhu, Shirong Ma, Peiyi Wang, Xiao Bi, et~al.
\newblock Deepseek-r1: Incentivizing reasoning capability in llms via reinforcement learning.
\newblock \emph{arXiv preprint arXiv:2501.12948}, 2025{\natexlab{a}}.

\bibitem[Guo et~al.(2025{\natexlab{b}})Guo, Zhang, Tong, Zhao, Huang, Zhang, Zhang, Liu, Zhang, Gao, et~al.]{guo2025can}
Ziyu Guo, Renrui Zhang, Chengzhuo Tong, Zhizheng Zhao, Rui Huang, Haoquan Zhang, Manyuan Zhang, Jiaming Liu, Shanghang Zhang, Peng Gao, et~al.
\newblock Can we generate images with cot? let's verify and reinforce image generation step by step.
\newblock \emph{arXiv preprint arXiv:2501.13926}, 2025{\natexlab{b}}.

\bibitem[Hao et~al.(2024)Hao, Romero, Lin, and Liu]{hao2024meshtron}
Zekun Hao, David~W Romero, Tsung-Yi Lin, and Ming-Yu Liu.
\newblock Meshtron: High-fidelity, artist-like 3d mesh generation at scale.
\newblock \emph{arXiv preprint arXiv:2412.09548}, 2024.

\bibitem[Huang et~al.(2025)Huang, Jia, Zhai, Cao, Ye, Zhao, Xu, Hu, and Lin]{huang2025vision}
Wenxuan Huang, Bohan Jia, Zijie Zhai, Shaosheng Cao, Zheyu Ye, Fei Zhao, Zhe Xu, Yao Hu, and Shaohui Lin.
\newblock Vision-r1: Incentivizing reasoning capability in multimodal large language models.
\newblock \emph{arXiv preprint arXiv:2503.06749}, 2025.

\bibitem[Jiang et~al.(2025)Jiang, Guo, Zhang, Zong, Li, Zhuo, Yan, Heng, and Li]{jiang2025t2i}
Dongzhi Jiang, Ziyu Guo, Renrui Zhang, Zhuofan Zong, Hao Li, Le Zhuo, Shilin Yan, Pheng-Ann Heng, and Hongsheng Li.
\newblock T2i-r1: Reinforcing image generation with collaborative semantic-level and token-level cot.
\newblock \emph{arXiv preprint arXiv:2505.00703}, 2025.

\bibitem[Khanna et~al.(2024)Khanna, Mao, Jiang, Haresh, Shacklett, Batra, Clegg, Undersander, Chang, and Savva]{khanna2024habitat}
Mukul Khanna, Yongsen Mao, Hanxiao Jiang, Sanjay Haresh, Brennan Shacklett, Dhruv Batra, Alexander Clegg, Eric Undersander, Angel~X Chang, and Manolis Savva.
\newblock Habitat synthetic scenes dataset (hssd-200): An analysis of 3d scene scale and realism tradeoffs for objectgoal navigation.
\newblock In \emph{Proceedings of the IEEE/CVF Conference on Computer Vision and Pattern Recognition}, pages 16384--16393, 2024.

\bibitem[Li et~al.(2024)Li, Zhang, Guo, Zhang, Li, Zhang, Zhang, Zhang, Li, Liu, et~al.]{li2024llava}
Bo Li, Yuanhan Zhang, Dong Guo, Renrui Zhang, Feng Li, Hao Zhang, Kaichen Zhang, Peiyuan Zhang, Yanwei Li, Ziwei Liu, et~al.
\newblock Llava-onevision: Easy visual task transfer.
\newblock \emph{arXiv preprint arXiv:2408.03326}, 2024.

\bibitem[Lin et~al.(2024)Lin, Ye, Zhu, Cui, Ning, Jin, and Yuan]{lin2024video}
Bin Lin, Yang Ye, Bin Zhu, Jiaxi Cui, Munan Ning, Peng Jin, and Li Yuan.
\newblock Video-llava: Learning united visual representation by alignment before projection.
\newblock In \emph{Proceedings of the 2024 Conference on Empirical Methods in Natural Language Processing}, pages 5971--5984, 2024.

\bibitem[Liu et~al.(2025)Liu, Liu, Liang, Li, Liu, Wang, Wan, Zhang, and Ouyang]{liu2025flow}
Jie Liu, Gongye Liu, Jiajun Liang, Yangguang Li, Jiaheng Liu, Xintao Wang, Pengfei Wan, Di Zhang, and Wanli Ouyang.
\newblock Flow-grpo: Training flow matching models via online rl.
\newblock \emph{arXiv preprint arXiv:2505.05470}, 2025.

\bibitem[o3 and o4~mini(https://openai.com/o3/)]{openai2025}
OpenAI: Introducing~OpenAI o3 and o4 mini.
\newblock 2025.
\newblock \emph{(2025)}, https://openai.com/o3/.

\bibitem[Qi et~al.(2024)Qi, Dong, Zhang, Geng, Han, Ge, Yi, and Ma]{qi2024shapellm}
Zekun Qi, Runpei Dong, Shaochen Zhang, Haoran Geng, Chunrui Han, Zheng Ge, Li Yi, and Kaisheng Ma.
\newblock Shapellm: Universal 3d object understanding for embodied interaction.
\newblock In \emph{European Conference on Computer Vision}, pages 214--238. Springer, 2024.

\bibitem[Rafailov et~al.(2023)Rafailov, Sharma, Mitchell, Manning, Ermon, and Finn]{rafailov2023direct}
Rafael Rafailov, Archit Sharma, Eric Mitchell, Christopher~D Manning, Stefano Ermon, and Chelsea Finn.
\newblock Direct preference optimization: Your language model is secretly a reward model.
\newblock \emph{Advances in neural information processing systems}, 36:\penalty0 53728--53741, 2023.

\bibitem[Schulman et~al.(2017)Schulman, Wolski, Dhariwal, Radford, and Klimov]{schulman2017proximal}
John Schulman, Filip Wolski, Prafulla Dhariwal, Alec Radford, and Oleg Klimov.
\newblock Proximal policy optimization algorithms.
\newblock \emph{arXiv preprint arXiv:1707.06347}, 2017.

\bibitem[Seed et~al.(2025)Seed, Zhang, Su, Sun, Xi, Xiao, Zheng, Zhang, Liu, Zan, et~al.]{seed2025seed}
ByteDance Seed, Yuyu Zhang, Jing Su, Yifan Sun, Chenguang Xi, Xia Xiao, Shen Zheng, Anxiang Zhang, Kaibo Liu, Daoguang Zan, et~al.
\newblock Seed-coder: Let the code model curate data for itself.
\newblock \emph{arXiv preprint arXiv:2506.03524}, 2025.

\bibitem[Shao et~al.(2024)Shao, Wang, Zhu, Xu, Song, Bi, Zhang, Zhang, Li, Wu, et~al.]{shao2024deepseekmath}
Zhihong Shao, Peiyi Wang, Qihao Zhu, Runxin Xu, Junxiao Song, Xiao Bi, Haowei Zhang, Mingchuan Zhang, YK Li, Yang Wu, et~al.
\newblock Deepseekmath: Pushing the limits of mathematical reasoning in open language models.
\newblock \emph{arXiv preprint arXiv:2402.03300}, 2024.

\bibitem[Shen et~al.(2025)Shen, Liu, Li, Fang, Ma, Liao, Shen, Zhang, Zhao, Zhang, et~al.]{shen2025vlm}
Haozhan Shen, Peng Liu, Jingcheng Li, Chunxin Fang, Yibo Ma, Jiajia Liao, Qiaoli Shen, Zilun Zhang, Kangjia Zhao, Qianqian Zhang, et~al.
\newblock Vlm-r1: A stable and generalizable r1-style large vision-language model.
\newblock \emph{arXiv preprint arXiv:2504.07615}, 2025.

\bibitem[Siddiqui et~al.(2024)Siddiqui, Alliegro, Artemov, Tommasi, Sirigatti, Rosov, Dai, and Nie{\ss}ner]{siddiqui2024meshgpt}
Yawar Siddiqui, Antonio Alliegro, Alexey Artemov, Tatiana Tommasi, Daniele Sirigatti, Vladislav Rosov, Angela Dai, and Matthias Nie{\ss}ner.
\newblock Meshgpt: Generating triangle meshes with decoder-only transformers.
\newblock In \emph{Proceedings of the IEEE/CVF conference on computer vision and pattern recognition}, pages 19615--19625, 2024.

\bibitem[Stojanov et~al.(2021)Stojanov, Thai, and Rehg]{stojanov2021using}
Stefan Stojanov, Anh Thai, and James~M Rehg.
\newblock Using shape to categorize: Low-shot learning with an explicit shape bias.
\newblock In \emph{Proceedings of the IEEE/CVF conference on computer vision and pattern recognition}, pages 1798--1808, 2021.

\bibitem[Tang et~al.(2024)Tang, Chen, Chen, Wang, Zeng, and Liu]{tang2024lgm}
Jiaxiang Tang, Zhaoxi Chen, Xiaokang Chen, Tengfei Wang, Gang Zeng, and Ziwei Liu.
\newblock Lgm: Large multi-view gaussian model for high-resolution 3d content creation.
\newblock In \emph{European Conference on Computer Vision}, pages 1--18. Springer, 2024.

\bibitem[Tong et~al.(2025)Tong, Guo, Zhang, Shan, Wei, Xing, Li, and Heng]{tong2025delving}
Chengzhuo Tong, Ziyu Guo, Renrui Zhang, Wenyu Shan, Xinyu Wei, Zhenghao Xing, Hongsheng Li, and Pheng-Ann Heng.
\newblock Delving into rl for image generation with cot: A study on dpo vs. grpo.
\newblock \emph{arXiv preprint arXiv:2505.17017}, 2025.

\bibitem[Wang et~al.(2025)Wang, Li, Zang, Wang, Lu, Jin, and Wang]{wang2025unified}
Yibin Wang, Zhimin Li, Yuhang Zang, Chunyu Wang, Qinglin Lu, Cheng Jin, and Jiaqi Wang.
\newblock Unified multimodal chain-of-thought reward model through reinforcement fine-tuning.
\newblock \emph{arXiv preprint arXiv:2505.03318}, 2025.

\bibitem[Wang et~al.(2024{\natexlab{a}})Wang, Lorraine, Wang, Su, Zhu, Fidler, and Zeng]{wang2024llama}
Zhengyi Wang, Jonathan Lorraine, Yikai Wang, Hang Su, Jun Zhu, Sanja Fidler, and Xiaohui Zeng.
\newblock Llama-mesh: Unifying 3d mesh generation with language models.
\newblock \emph{arXiv preprint arXiv:2411.09595}, 2024{\natexlab{a}}.

\bibitem[Wang et~al.(2024{\natexlab{b}})Wang, Wang, Chen, Xiang, Chen, Yu, Li, Su, and Zhu]{wang2024crm}
Zhengyi Wang, Yikai Wang, Yifei Chen, Chendong Xiang, Shuo Chen, Dajiang Yu, Chongxuan Li, Hang Su, and Jun Zhu.
\newblock Crm: Single image to 3d textured mesh with convolutional reconstruction model.
\newblock In \emph{European conference on computer vision}, pages 57--74. Springer, 2024{\natexlab{b}}.

\bibitem[Wu et~al.(2025)Wu, Lin, Zhang, Zeng, Yang, Bao, Qian, Zhu, Cao, Torr, et~al.]{wu2025direct3d}
Shuang Wu, Youtian Lin, Feihu Zhang, Yifei Zeng, Yikang Yang, Yajie Bao, Jiachen Qian, Siyu Zhu, Xun Cao, Philip Torr, et~al.
\newblock Direct3d-s2: Gigascale 3d generation made easy with spatial sparse attention.
\newblock \emph{arXiv preprint arXiv:2505.17412}, 2025.

\bibitem[Wu et~al.(2023)Wu, Hao, Sun, Chen, Zhu, Zhao, and Li]{wu2023human}
Xiaoshi Wu, Yiming Hao, Keqiang Sun, Yixiong Chen, Feng Zhu, Rui Zhao, and Hongsheng Li.
\newblock Human preference score v2: A solid benchmark for evaluating human preferences of text-to-image synthesis.
\newblock \emph{arXiv preprint arXiv:2306.09341}, 2023.

\bibitem[Xiang et~al.(2025)Xiang, Lv, Xu, Deng, Wang, Zhang, Chen, Tong, and Yang]{xiang2025structured}
Jianfeng Xiang, Zelong Lv, Sicheng Xu, Yu Deng, Ruicheng Wang, Bowen Zhang, Dong Chen, Xin Tong, and Jiaolong Yang.
\newblock Structured 3d latents for scalable and versatile 3d generation.
\newblock In \emph{Proceedings of the Computer Vision and Pattern Recognition Conference}, pages 21469--21480, 2025.

\bibitem[Xu et~al.(2023)Xu, Wang, Cheng, Cao, Shan, Qie, and Gao]{xu2023dream3d}
Jiale Xu, Xintao Wang, Weihao Cheng, Yan-Pei Cao, Ying Shan, Xiaohu Qie, and Shenghua Gao.
\newblock Dream3d: Zero-shot text-to-3d synthesis using 3d shape prior and text-to-image diffusion models.
\newblock In \emph{Proceedings of the IEEE/CVF Conference on Computer Vision and Pattern Recognition}, pages 20908--20918, 2023.

\bibitem[Xue et~al.(2025)Xue, Wu, Gao, Kong, Zhu, Chen, Liu, Liu, Guo, Huang, et~al.]{xue2025dancegrpo}
Zeyue Xue, Jie Wu, Yu Gao, Fangyuan Kong, Lingting Zhu, Mengzhao Chen, Zhiheng Liu, Wei Liu, Qiushan Guo, Weilin Huang, et~al.
\newblock Dancegrpo: Unleashing grpo on visual generation.
\newblock \emph{arXiv preprint arXiv:2505.07818}, 2025.

\bibitem[Yang et~al.(2025)Yang, Li, Yang, Zhang, Hui, Zheng, Yu, Gao, Huang, Lv, et~al.]{yang2025qwen3}
An Yang, Anfeng Li, Baosong Yang, Beichen Zhang, Binyuan Hui, Bo Zheng, Bowen Yu, Chang Gao, Chengen Huang, Chenxu Lv, et~al.
\newblock Qwen3 technical report.
\newblock \emph{arXiv preprint arXiv:2505.09388}, 2025.

\bibitem[Yang et~al.(2024)Yang, Shi, Zhang, Yang, Wang, Zhao, Liu, Wang, Lin, Yu, et~al.]{yang2024hunyuan3d}
Xianghui Yang, Huiwen Shi, Bowen Zhang, Fan Yang, Jiacheng Wang, Hongxu Zhao, Xinhai Liu, Xinzhou Wang, Qingxiang Lin, Jiaao Yu, et~al.
\newblock Hunyuan3d 1.0: A unified framework for text-to-3d and image-to-3d generation.
\newblock \emph{arXiv preprint arXiv:2411.02293}, 2024.

\bibitem[Ye et~al.(2025)Ye, Wang, Zhao, Xie, and Zhu]{ye2025shapellm}
Junliang Ye, Zhengyi Wang, Ruowen Zhao, Shenghao Xie, and Jun Zhu.
\newblock Shapellm-omni: A native multimodal llm for 3d generation and understanding.
\newblock \emph{arXiv preprint arXiv:2506.01853}, 2025.

\bibitem[Yu et~al.(2025)Yu, Zhang, Zhu, Yuan, Zuo, Yue, Dai, Fan, Liu, Liu, et~al.]{yu2025dapo}
Qiying Yu, Zheng Zhang, Ruofei Zhu, Yufeng Yuan, Xiaochen Zuo, Yu Yue, Weinan Dai, Tiantian Fan, Gaohong Liu, Lingjun Liu, et~al.
\newblock Dapo: An open-source llm reinforcement learning system at scale.
\newblock \emph{arXiv preprint arXiv:2503.14476}, 2025.

\bibitem[Zhang et~al.(2024)Zhang, Jiang, Zhang, Lin, Guo, Qiu, Zhou, Lu, Chang, Qiao, et~al.]{zhang2024mathverse}
Renrui Zhang, Dongzhi Jiang, Yichi Zhang, Haokun Lin, Ziyu Guo, Pengshuo Qiu, Aojun Zhou, Pan Lu, Kai-Wei Chang, Yu Qiao, et~al.
\newblock Mathverse: Does your multi-modal llm truly see the diagrams in visual math problems?
\newblock In \emph{European Conference on Computer Vision}, pages 169--186. Springer, 2024.

\bibitem[Zhao et~al.(2025{\natexlab{a}})Zhao, Ye, Wang, Liu, Chen, Wang, and Zhu]{zhao2025deepmesh}
Ruowen Zhao, Junliang Ye, Zhengyi Wang, Guangce Liu, Yiwen Chen, Yikai Wang, and Jun Zhu.
\newblock Deepmesh: Auto-regressive artist-mesh creation with reinforcement learning.
\newblock In \emph{Proceedings of the IEEE/CVF International Conference on Computer Vision}, pages 10612--10623, 2025{\natexlab{a}}.

\bibitem[Zhao et~al.(2025{\natexlab{b}})Zhao, Lai, Lin, Zhao, Liu, Yang, Feng, Yang, Zhang, Yang, et~al.]{zhao2025hunyuan3d}
Zibo Zhao, Zeqiang Lai, Qingxiang Lin, Yunfei Zhao, Haolin Liu, Shuhui Yang, Yifei Feng, Mingxin Yang, Sheng Zhang, Xianghui Yang, et~al.
\newblock Hunyuan3d 2.0: Scaling diffusion models for high resolution textured 3d assets generation.
\newblock \emph{arXiv preprint arXiv:2501.12202}, 2025{\natexlab{b}}.

\bibitem[Zheng et~al.(2025{\natexlab{a}})Zheng, Liu, Li, Chen, Yu, Gao, Dang, Liu, Men, Yang, et~al.]{zheng2025group}
Chujie Zheng, Shixuan Liu, Mingze Li, Xiong-Hui Chen, Bowen Yu, Chang Gao, Kai Dang, Yuqiong Liu, Rui Men, An Yang, et~al.
\newblock Group sequence policy optimization.
\newblock \emph{arXiv preprint arXiv:2507.18071}, 2025{\natexlab{a}}.

\bibitem[Zheng et~al.(2025{\natexlab{b}})Zheng, Yang, Hong, Zhao, Xu, Yang, Shen, and Yu]{zheng2025deepeyes}
Ziwei Zheng, Michael Yang, Jack Hong, Chenxiao Zhao, Guohai Xu, Le Yang, Chao Shen, and Xing Yu.
\newblock Deepeyes: Incentivizing" thinking with images" via reinforcement learning.
\newblock \emph{arXiv preprint arXiv:2505.14362}, 2025{\natexlab{b}}.

\end{thebibliography}
